\documentclass[11pt]{article}

\usepackage[preprint]{acl}

\usepackage{times}
\usepackage{latexsym}

\usepackage[T1]{fontenc}

\usepackage[utf8]{inputenc}

\usepackage{authblk}
\usepackage{amsthm}    
\usepackage[utf8]{inputenc} 
\usepackage[T1]{fontenc} 
\usepackage{microtype} 
\usepackage{amsmath,amssymb,amsfonts} 
\usepackage{nicefrac} 
\usepackage{graphicx} 
\usepackage{textcomp} 
\usepackage{svg} 
\usepackage{lipsum} 
\usepackage[table]{xcolor} 
\usepackage{booktabs} 
\usepackage{multirow} 
\usepackage{makecell} 
\usepackage{siunitx} 
\usepackage{caption} 
\usepackage{subcaption} 
\usepackage{algorithm} 
\usepackage{algpseudocode} 
\usepackage{geometry} 
\usepackage{url} 
\usepackage{hyperref} 
\usepackage[ruled,linesnumbered,algo2e,vlined]{algorithm2e}
\usepackage{amsmath}
\usepackage{tcolorbox}
\usepackage{xcolor}  
\usepackage{wrapfig}
\usepackage{amssymb}
\usepackage{mathtools}
\usepackage{xcolor}
\usepackage{colortbl} 
\definecolor{ImproveGreen}{RGB}{0,150,80} 
\definecolor{ReduceBlue}{RGB}{0,90,180} 

\newtheorem{proposition}{Proposition} %
\newtheorem{lemma}{Lemma} 
\newtheorem{theorem}{Theorem} %
\newtheorem{definition}{Definition}
\usepackage{graphicx}

\author{Guangya Wan$^{*}$, Zixin Stephen Xu$^{*}$, Sasa Zorc, Manel Baucells,  \\ Mengxuan Hu, Hao Wang, Sheng Li$^{\dagger}$ \\
University of Virginia \\
\small $^{*}$Equal contribution \quad $^{\dagger}$Corresponding author}

\begin{document}

\title{BEACON: Bayesian Optimal Stopping for Efficient LLM Sampling}
\maketitle
\begin{abstract} 
Sampling multiple responses is a common way to improve LLM output quality, but it comes at the cost of additional computation. The key challenge is deciding when to stop generating new samples to balance accuracy gains against efficiency. To address this, we introduce BEACON (Bayesian Efficient Adaptive Criterion for Optimal N-stopping), a principled adaptive sampling framework grounded in Sequential Search with Bayesian Learning. BEACON sequentially generates responses from the policy LLM, updates posterior belief over reward distributions in real time without further training, and determines when to stop by weighing expected gains against computational cost. Sampling terminates once the marginal utility of further exploration no longer justifies the expense. We establish both theoretical optimality guarantees and practical tractability, and show empirically that BEACON reduces average sampling by up to 80\% while maintaining response quality. We further demonstrate BEACON’s utility for cost-efficient preference data generation and outline practical extensions, offering actionable insights for future researchers.

\end{abstract}

\section{Introduction}


\begin{figure*}[hbt!] 
  \centering
  \includegraphics[width=0.88\textwidth]{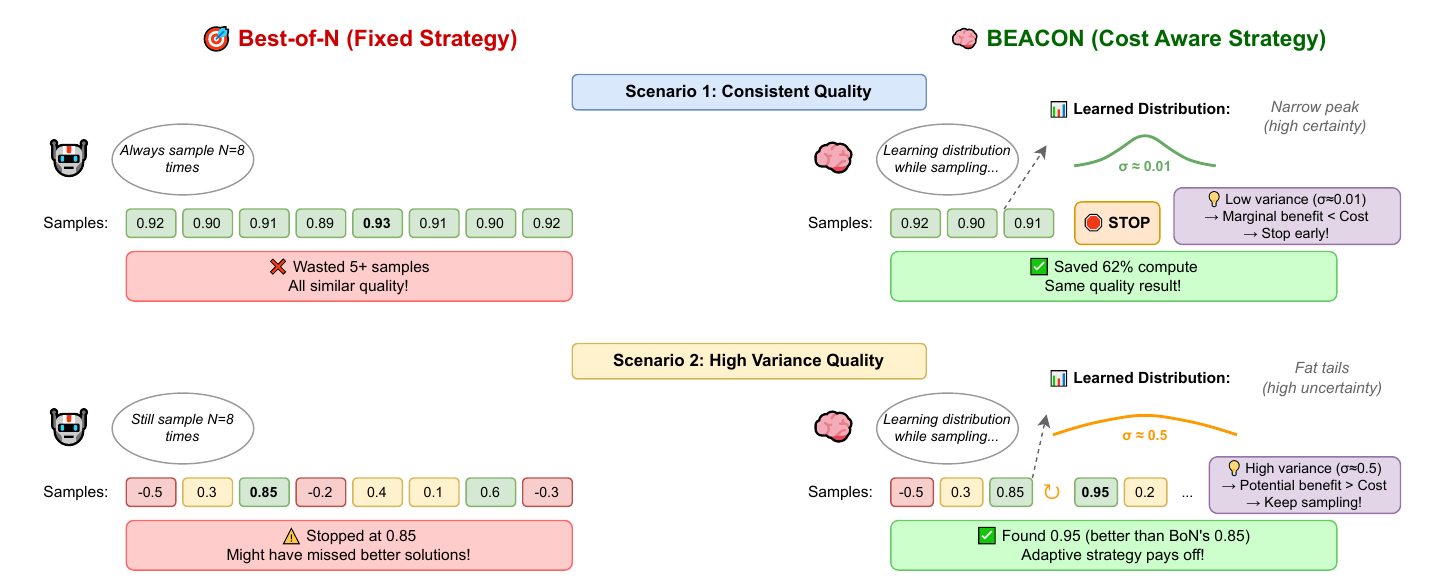}
\caption{\textbf{Comparison of BEACON adaptive sampling versus Best-of-N fixed sampling.} BEACON adaptively determines sample size by learning the reward distribution and determine if additional sampling is worth the cost. Intuitively, BEACON stops earlier for consistent samples and continues sampling to find better solutions for variable reward samples, while Best-of-N always uses fixed samples (in this case, 8) regardless.}. 
  \label{fig:bas_comparison}
  \vspace{-12pt}
\end{figure*}


Large Language Models (LLMs) have shown human-like abilities across diverse tasks such as mathematics, coding, and creative writing \citep{ke2025surveyfrontiersllmreasoning,hendrycksmath2021}. Yet, they often produce inconsistent outputs, occasionally hallucinated on queries they could solve correctly across different runs \citep{manakul2023selfcheckgpt,xu2025hallucinationinevitableinnatelimitation}. To address this, \textbf{sampling} has been widely adopted: by generating multiple responses and selecting one based on specific criteria, it improves performance in tasks like complex reasoning \citep{wang2022self,snell2025scaling}, safety alignment \citep{ichihara2025evaluation}, and preference data generation \citep{yuan2024scaling}. However, blindly scaling computational resources is suboptimal and impractical, particularly in settings such as streaming or real-time LLM applications \citep{xiao2024efficient}, where efficiency is as critical as \textbf{response quality} \citep{yehudai2025surveyevaluationllmbasedagents}. This highlights the need for a deeper understanding of the \textbf{economy of inference}—balancing computational cost against performance gains.

Existing adaptive sampling methods are mainly based on sample-consistency heuristics to estimate task difficulty or confidence \citep{aggarwal-etal-2023-lets,wang2022self,wan2025derailerreraileradaptiveverificationefficient,taubenfeld2025confidenceimprovesselfconsistencyllms,wan-etal-2025-reasoning}. While training-free and easy to implement, these approaches often fail to generalize \citep{Fu2024Certaindex,wang-etal-2025-make} because multiple incorrect responses can exhibit consistency, and measuring consistency remains challenging for \textbf{open-ended tasks} with multiple valid answers. An alternative direction focuses on making Best-of-N sampling adaptive by learning when to stop generating candidates based on \textbf{reward model feedback} \citep{cobbe2021training,OpenAI2022,zhang2024generative}. While these adaptive BoN methods show effectiveness across diverse scenarios, they rely on data-centric, training-heavy pipelines to learn auxiliary stopping models \citep{damani2025learning}, which limits adaptability to new domains while potentially introducing bias and reducing output diversity. Critically, both approaches rely on heuristics or learned approximations without theoretical guarantees of optimality, making their stopping decisions inherently ad-hoc.

 To bridge this theory-practice gap, we leverage principles in Bayesian optimal stopping \citep{baucells2024search} and reformulate LLM sampling as a sequential search problem. This framework ensures stopping decisions achieve \textbf{Bayesian optimality} given currently observed data, eliminating reliance on heuristic approximations \citep{rothschild1974searching}. Rather than learning reward distributions \emph{offline}, we conceptualize them as latent processes for online updating: each generated response reveals information about the underlying reward distribution while incurring computational costs \citep{pmlr-v119-toth20a}. The fundamental challenge becomes determining the \textbf{optimal stopping point} where expected benefits from additional samples no longer justify associated costs, which can be addressed with Bayesian learning theory \citep{christensen1986,bikhchandani1996optimal}. 

We therefore introduce the \textbf{Bayesian Efficient Adaptive Criterion for Optimal N-stopping (BEACON)}, 
a novel adaptive sampling framework that makes optimal stopping decisions computationally 
practical while enabling real-time deployment without additional offline training requirements. 
Our approach can be understood through two synergistic components: \textbf{sequential search} addresses 
the adaptivity challenge, while \textbf{Bayesian learning} provides a principled framework for online 
reward distribution learning. Together, these components enable derivation of adaptive sampling policies 
without pre-training while guaranteeing \textbf{theoretical optimality}. 
BEACON sequentially collects responses, updates sufficient statistics of the posterior reward distribution, 
and employs an index-based sampling policy that compares a quality index against a cost threshold. 
Intuitively, BEACON terminates when reward evaluations exhibit minimal variation, indicating stable 
characterization of the quality distribution, or when additional computation is unlikely to yield 
superior rewards. Figure~\ref{fig:bas_comparison} contrasts BEACON’s adaptive stopping with conventional 
Best-of-N sampling, illustrating how our Bayesian criterion adaptively allocates computation across 
variable-reward queries. Our empirical evaluations on reasoning and alignment benchmarks demonstrate 
that BEACON substantially reduces average inference costs compared to fixed BoN while maintaining 
comparable performance, with demonstrated utility for cost-effective preference data generation, 
practical hyperparameter selection guidance, and extensions to batch sampling for enhanced efficiency. 
In sum, our main contributions are: 
(\textbf{1}) We propose BEACON, an \textbf{adaptive sampling framework} that reformulates LLM sampling 
as a sequential Bayesian search problem for theoretically optimal stopping without training additional 
auxiliary models. 
(\textbf{2}) We provide rigorous analysis of its \textbf{theoretical guarantees} and computational 
complexity. 
(\textbf{3}) We conduct comprehensive experiments across diverse benchmarks, demonstrating its 
effectiveness against established baselines and highlighting practical extensions for 
post-training and real-world deployment.

\begin{figure*}[t] 
  \centering
  \includegraphics[width=0.8\textwidth]{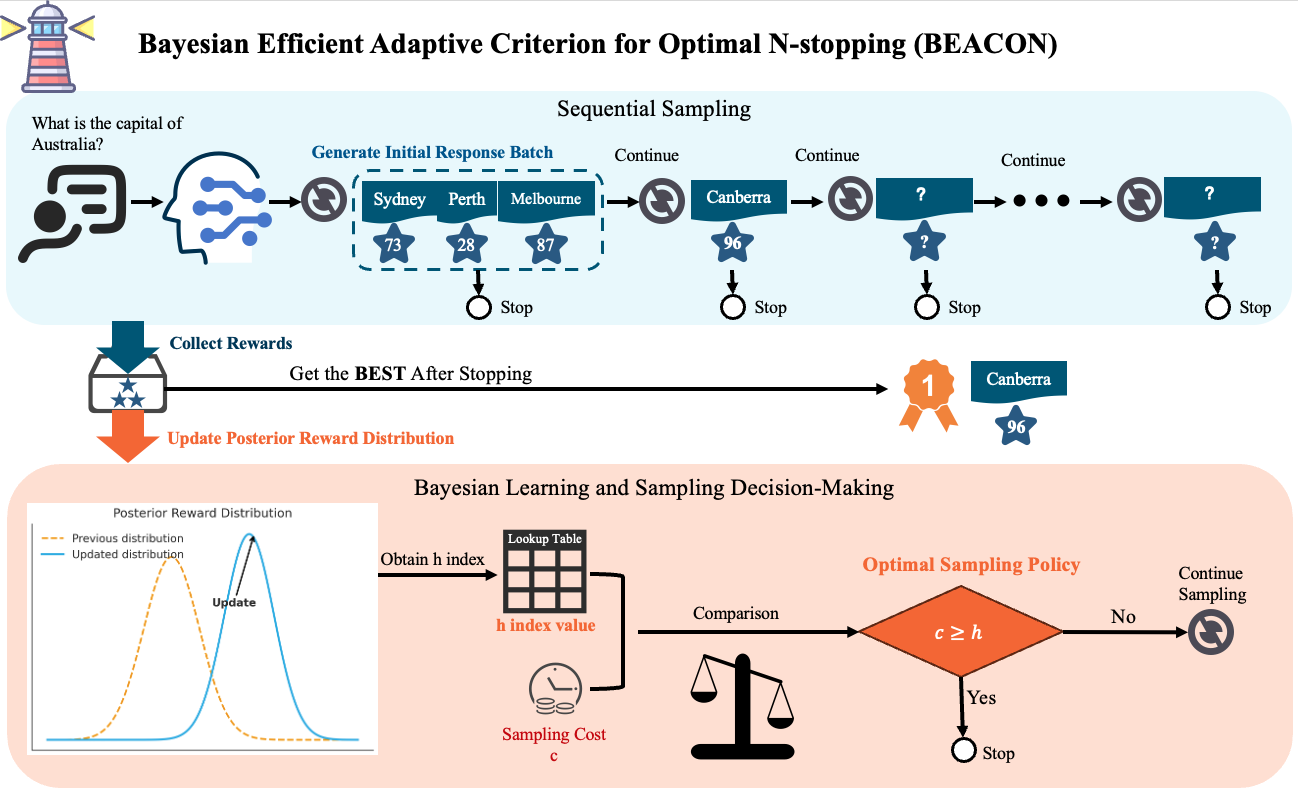}
  \caption{\textbf{BEACON framework}: The top layer shows sequential sampling of LLM responses with reward model evaluation. The bottom layer illustrates the optimal stopping mechanism, which updates Bayesian posterior beliefs about reward distribution parameters after each sample and determines when to stop based on optimal sampling policy, comparing the index-based threshold to the sampling cost.}
  \label{fig:bas_pipeline}
  \vspace{-8pt}
\end{figure*}

\section{Modeling LLM Sampling as Sequential Search Problem}
\subsection{Methodology Overview}
\noindent \textbf{Reward Models as Quality Assessment Tools.} Reward Models (RMs) \citep{zhang2024generative, zhong2025comprehensivesurveyrewardmodels} provide scalar assessments of response quality that serve as evaluation signals for adaptive sampling. Trained on pairwise preference data $D = \{(x_i, y_{w,i}, y_{l,i})\}$, these models encode both preference direction and certainty in the magnitude of reward differences, making them ideal quality signals for probabilistic modeling.

\noindent \textbf{Sequential Search for Optimal Stopping.} We reformulate LLM sampling as a sequential search problem to maximize net gain—balancing the highest quality against sampling cost. This approach replaces heuristics with theoretically grounded guarantees for deciding when additional samples are no longer economically justified. Sequential Search examines alternatives one by one, deciding after each observation whether to accept the current best outcome or continue sampling \citep{stigler1961economics, weitzman1979optimal} (detailed in Appendix \ref{app:sequential_search}). Given observed rewards $\mathbf{r}_k = \{r_1, \dots, r_k\}$ with best reward $z_k = \max\{r_1, \dots, r_k\}$ and sampling cost $c$ per observation, the challenge is determining the optimal stopping point, identifying the maximum reward while minimizing costs. When the reward distribution is \textit{known}, this admits closed-form solutions \citep{weitzman1979optimal}. However, LLM sampling presents the more challenging case where the underlying reward distribution is \textit{unknown}.

\noindent \textbf{A Principled Bayesian Framework for Unknown Distributions.} BEACON combines sequential search with Bayesian learning to address the fundamental challenge of \textit{unknown reward distributions} by learning parameters online during sampling, enabling zero-shot deployment without offline training or pre-training. For computational tractability, we employ conjugate priors that enable closed-form updates. We focus on the Normal distribution for its practical utility and unique theoretical properties—it is the only continuous distribution with computationally efficient optimal index policies in sequential search literature \citep{baucells2024search}, which is more challenging than simpler, discrete conjugate families such as the beta-binomial extension that is also supported by BEACON (demonstrated in Appendix~\ref{app:discrete_case}).


\RestyleAlgo{ruled}

\SetCommentSty{textcolor{gray}}
\begin{algorithm*}[t]
\caption{BEACON: Bayesian Efficient Adaptive Criterion for Optimal N-stopping}
\label{alg:beacon}
\DontPrintSemicolon
\SetAlgoLined
\SetKwInput{KwInput}{Input}
\SetKwInput{KwOutput}{Output}

\KwInput{Query $x$, policy LLM $\pi_\phi(y|x)$, reward model $R(x,y)$, cost $c$, max samples $n$, grid size $G$}
\KwOutput{Best response $y^*$ and its reward $r^*$}

\BlankLine
\tcp{Step 1: Initialize prior and h-index table}
$(\alpha_0, \nu_0, \beta_0, \mu_0) \gets (-0.5, 0, 0, 0)$ \Comment{Non-informative prior}

$h\text{-}table \gets \text{PrecomputeTable}(n,G,\alpha_0,\nu_0)$ \Comment{Pre-compute as in \S\ref{app:compute_h_index}}

\BlankLine
\tcp{Step 2: Generate initial samples and compute baseline parameters}
Generate $\{y_1, y_2, y_3\} \sim \pi_\phi(y|x)$ and compute $r_i \gets R(x, y_i)$ for $i \in \{1,2,3\}$

$(\alpha,\nu, \mu,\beta) \gets \text{UpdateHyperParams}(r_1, r_2, r_3, \alpha_0, \nu_0, \beta_0, \mu_0)$ 

$z \gets \max\{r_1,r_2,r_3\}$, $\sigma \gets \sqrt{\tfrac{(\nu+1)\beta}{\nu\alpha}}$, $k \gets 3$

\BlankLine
\tcp{Step 3: Adaptive sampling loop}
\While{$k < n$}{
    $\hat{z} \gets (z - \mu)/\sigma$,
    $h \gets \text{LookupHIndex}(h\text{-}table, k, \hat{z})$ 
    
    \If{$h \leq c/\sigma$}{
        \textbf{break} \Comment{Apply UIP as in \eqref{eq:UIP}}
    }
    
    Generate $y_k \sim \pi_\phi(y|x)$ and compute $r_k \gets R(x, y_k)$
    $k \gets k + 1$
    
    $q_{0.01} \gets F_{2\alpha_{k-1}}^{-1}(0.01| \mu, \sigma)$
    $\tilde{r}_k \gets 
    \begin{cases} 
        \mu & \text{if } r_k < q_{0.01} \\
        r_k & \text{otherwise}
    \end{cases}$ \Comment{Filter extreme low values}
    
    $(z,\mu, \sigma) \gets \text{UpdateStats}(z, \mu, \sigma, \tilde{r}_k, k)$ \Comment{The sufficient statistics updated by \eqref{parameter_update}}} 

\BlankLine
$i^* \gets \arg\max_{i\in\{1,\ldots,k\}} r_i$
\Return{$y_{i^*}$, $r_{i^*}$}
\end{algorithm*}

\subsection{Model Setup}\label{sec:model}
\noindent \textbf{Problem Setting.} Given a query $x$ and a policy LLM $\pi_\phi(y|x)$, we sequentially generate responses $\{y_1, \ldots, y_k\}$ where $y_i \sim \pi_\phi(\cdot|x)$ and evaluate each using a reward model $R(x, y_i) = r_i$. The reward distribution $f(r_k|x)$ emerges from the marginalized distribution over the response generation and reward evaluation processes. We assume rewards follow an i.i.d. Normal distribution with unknown parameters, which distinguishes our approach from methods requiring known distributions or training auxiliary models. After collecting $k$ samples with rewards $\mathbf{r}_k = \{r_1, \ldots, r_k\}$, we denote the current best reward as $z_k = \max\{r_1, \ldots, r_k\}$. Each sample incurs cost $c$, and sampling continues up to maximum horizon $n$. Our objective is determining the optimal stopping time $K$ that maximizes expected net gain $\mathbb{E}[z_K - K \cdot c]$.

\noindent \textbf{Bayesian Learning with Conjugate Priors.} To enable tractable Bayesian updating, we employ a Normal-Inverse-Gamma (NIG) conjugate prior for unknown parameters $(\mu, \sigma^2)$. Conjugate priors guarantee a fixed-dimensional state space during sequential sampling \citep{diaconis1979}, essential for computational tractability. Starting with prior NIG$(\mu_0, \nu_0, \alpha_0, \beta_0)$, after observing $k \geq k_0$ samples (where $k_0$ is the minimum for well-defined posteriors; see Appendix \ref{app:minimal_sample_size}), the posterior parameters update according to closed-form expressions detailed as \eqref{eq:hyperamareters_update} in Appendix \ref{app:minimal_sample_size}. Importantly, the triple $(z_k, \mu_k, \sigma_k)$ forms \textbf{sufficient statistics} for all observed rewards, enabling efficient state representation where the updating formula \eqref{parameter_update} in Appendix \ref{app:sufficent_statistics} shows how these statistics evolve with each new sample.

\noindent \textbf{Bellman Equation.} To formulate our objective function $\mathbb{E}[z_K - K \cdot c]$ in Bellman equation as:
\begin{align}\label{eq:bellman_sufficient}
\begin{aligned}
&V_{n,k}(z_k,\mu_k,\sigma_k;c) =\\
&\max\{ z_k, \mathbb{E}[V_{n,k+1}(z_{k+1},\mu_{k+1},\sigma_{k+1};c)] - c \},
\end{aligned}
\end{align}
with corresponding optimal stopping rule:
\begin{align}\label{eq:optimal_stopping_specific}
\begin{aligned}
&\text{Stop iff}\quad H_{n,k}(z_k,\mu_k,\sigma_k;c) = \\
&\mathbb{E}[V_{n,k+1}(z_{k+1},\mu_{k+1},\sigma_{k+1};c)] - z_k \leq c.
\end{aligned}
\end{align}
where $H_{n,k}$ represents the expected marginal gain from continued sampling.

\label{method:analysis}

\subsection{Optimal Sampling Policy}
Building on recent theoretical advances of the computationally efficient Universal Index Policy \citep{baucells2024search}, we can establish an efficient criterion for optimal stopping decisions.

\begin{definition}\label{def_h_index} 
For $k_0\leq k< n$, the \textit{h-index function} $h_{n,k}:\mathbb{R}\to (0,\infty)$ maps each standardized best reward $\hat{z}\in \mathbb{R}$ to the unique value $c>0$ that solves the condition where the expected marginal gain from continuing equals the sampling cost.
\end{definition}

\begin{theorem}[Optimal Sampling Policy] \label{thm:optimal_policy}
After generating initial samples $\{y_1, y_2, \ldots, y_{k_0}\}$ to establish valid posterior parameters, the optimal Bayesian policy at each step $k \geq k_0$ is to continue sampling if and only if:
\begin{equation}\label{eq:UIP}
h_{n,k}\left(\frac{z_k-\mu_k}{\sigma_k}\right) > \frac{c}{\sigma_k}
\end{equation}
The stopping time $K = \min\{k \geq k_0 : h_{n,k}(\hat{z}_k) \leq c/\sigma_k\} \wedge n$ maximizes the expected net gain $\mathbb{E}[z_K - K \cdot c]$.
\end{theorem}

\noindent The proof of Theorem~\ref{thm:optimal_policy} is provided in the Appendix. Our approach standardizes the current best reward $\hat{z}_k=(z_k-\mu_k)/\sigma_k$, retrieves the corresponding h-index value, and compares it against the cost-adjusted threshold $c/\sigma_k$. The algorithm stops when this threshold is no longer exceeded, indicating that further sampling has become economically inefficient given our posterior beliefs.

\medskip
\noindent \textbf{Normality Assumption.} 
While exact optimality guarantees hold under normality, BEACON remains 
robust to moderate distributional violations through several mechanisms: 
(1) the Central Limit Theorem suggests that reward signals naturally approximate normality in practice (as shown in App. \ref{app:diversity}); (2) our focus on identifying  maximum rewards depends on the right tail of the distribution, 
making the framework less sensitive to left-tail deviations; and (3) we 
introduce a robust updating mechanism (see Section~\ref{sec:extensions}) 
that filters extreme low outliers while preserving high-quality samples, 
maintaining practical effectiveness when distributions exhibit negative 
skewness.

\medskip
\noindent \textbf{Sensitivity Analysis.} The optimal stopping time $K$ exhibits intuitive dependencies on key problem parameters. When sampling cost $c$ increases, exploration is often discouraged. If the current best reward $z_k$ substantially exceeds the posterior mean $\mu_k$ (large $\hat{z}_k$), the framework recognizes an exceptionally high-quality sample has likely been found and stops earlier. Conversely, greater posterior uncertainty (larger $\sigma_k$) encourages continued sampling through two mechanisms: by decreasing the normalized score $\hat{z}_k$ and lowering the effective cost threshold $\frac{c}{\sigma_k}$. Intuitively, when more exploration budget remains available (larger $n-k$), the algorithm tends to be more patient, balancing immediate rewards against future exploration potential (see Appendix~\ref{app: prop1} for proofs and formal statements).


%
\subsection{BEACON Framework}
\noindent\textbf{Hyperparameter Configuration.} BEACON requires three key hyperparameters that jointly define the optimization framework: (1) \textit{Prior Parameters} -- We use Jeffreys' non-informative prior $(\alpha_0, \nu_0, \mu_0, \beta_0) = (-0.5, 0, 0, 0)$ for task-agnostic deployment without domain-specific calibration, requiring $k_0=3$ initial samples for well-defined posteriors (Appendix~\ref{app:minimal_sample_size}); (2) \textit{Maximum Horizon ($n$)} -- The sampling budget follows standard Best-of-N configurations (e.g. $n \in \{8, 16, 32\}$), where larger horizons increase patience but raise h-index pre-computation costs (Appendix~\ref{app:compute_h_index}); (3) \textit{Sampling Cost ($c$)} -- Controls the quality-efficiency trade-off, with higher values favoring efficiency and lower values favoring quality.

\medskip
\noindent\textbf{Algorithm Implementation.} Algorithm \ref{alg:beacon} presents the complete BEACON procedure, with the overall framework illustrated in Figure~\ref{fig:bas_pipeline}. After initializing Jeffreys' non-informative priors and pre-computing h-index tables, we generate $k_0=3$ bootstrap samples to establish valid posterior parameters. The adaptive sampling loop then iteratively: (1) computes standardized score $\hat{z}_k=(z_k-\mu_k)/\sigma_k$; (2) retrieves h-index $h_{n,k}(\hat{z}_k)$ via table lookup; (3) applies optimal stopping criterion $h_{n,k}(\hat{z}_k) \leq c/\sigma_k$; and (4) if continuing, generates new samples and updates parameters using robust filtering. This design transforms computationally intensive Bellman optimization into efficient table lookups, enabling real-time deployment while maintaining theoretical optimality guarantees.

\medskip
\noindent\textbf{Computational Complexity.}
\label{met:comp}
BEACON's computational overhead consists of two distinct components with different scalability characteristics: (1) \textit{h-Table Pre-computation} -- Constructing the lookup table $h_{n,k}(\cdot)$ requires $\mathcal{O}(nG)$ operations for horizon $n$ and grid resolution $G$. This one-time cost is amortized across all queries that share the same horizon, making it negligible in multi-query deployments. When tasks involve multiple horizons $\{n_1, \ldots, n_J\}$, complexity grows only with the number of distinct horizon values rather than the total number of queries; (2) \textit{Sequential Inference} -- Each query entails sequential decision-making, where samples are generated one-by-one to update posterior beliefs. This inherent dependency limits within-query parallelization and can increase latency relative to batch-generation methods. Nevertheless, the cost can be partially mitigated through batch-parallel sampling (Section~\ref{exp:batch}).

\section{Experiments}
\label{sec:results}
\begin{table*}[t]
\centering
\caption{Comparison of BEACON with baseline sampling methods across different models and tasks. BEACO achieves a superior trade-off between Accuracy/Win Rate/Reward $\bar{\hat{z}}_K$ and efficiency (Avg Sample $\overline{K}$), measured by the implicitly optimized objective $\bar{\hat{V}}_K$. Upward arrows (↑) indicate percentage improvement in accuracy or win rate over CoT baselines; downward arrows (↓) show percentage reduction in samples compared to the maximum BoN sample size (32). 
\textbf{Reward models:} \textbf{N-RM} uses \textit{Llama-3.1-Nemotron-70B-Reward} \citep{wang2025helpsteerpreference}; \textbf{S-RM} uses \textit{Skywork-Llama-3.1-8B} \citep{liu2024skyworkrewardbagtricksreward}; Additional runs for statistical signifiance reported in Table \ref{tab:llama_significance}.}
\label{tab:model_comparison_no_siunitx_updated}
\tiny
\renewcommand{\arraystretch}{1.25}
\setlength{\tabcolsep}{4.5pt}
\begin{tabular}{@{}l l r r r r | r r r r@{}}
\toprule
 & & \multicolumn{4}{c}{\textbf{Reasoning Tasks (Avg. MATH/AIME/AMC)}} & \multicolumn{4}{c}{\textbf{Alignment Task (AlpacaEval 2.0)}} \\
\cmidrule(lr){3-6} \cmidrule(lr){7-10}
\textbf{Model} & \textbf{Method} & \makecell[r]{\textbf{Accuracy ↑}\\\textbf{(\%)}} & \makecell[r]{\textbf{Samples ↓}\\\textbf{($\overline{K}$)}} & \makecell[r]{\textbf{Reward ↑}\\\textbf{($\bar{\hat{z}}_K$)}} & \makecell[r]{\textbf{Value ↑}\\\textbf{($\bar{\hat{V}}_K$)}} & \makecell[r]{\textbf{Win Rate ↑}\\\textbf{(\%)}} & \makecell[r]{\textbf{Samples ↓}\\\textbf{($\overline{K}$)}} & \makecell[r]{\textbf{Reward ↑}\\\textbf{($\bar{\hat{z}}_K$)}} & \makecell[r]{\textbf{Value↑}\\\textbf{($\bar{\hat{V}}_K$)}} \\
\midrule
\multirow{8}{*}{\texttt{LLaMA-3.2-3B}}
& Direct CoT     & 20.0 &  1.0 & -1.15 & -0.40 & 16.0 &  1.0 & -1.08 & -0.80 \\
& SC                & 28.5 \textcolor{ImproveGreen}{↑42.5\%} & 16.0 \textcolor{ReduceBlue}{↓50.0\%} & -0.35 & -0.01 &    - &    - &    - &    - \\
& RASC              & 27.8 \textcolor{ImproveGreen}{↑39.0\%} &  5.2 \textcolor{ReduceBlue}{↓83.8\%} & -0.30 & 0.66 & 20.0 \textcolor{ImproveGreen}{↑25.0\%} &  7.0 \textcolor{ReduceBlue}{↓78.1\%} & -1.12 & -0.55 \\
& AS                & 27.8 \textcolor{ImproveGreen}{↑39.0\%} &  5.6 \textcolor{ReduceBlue}{↓82.5\%} & -0.31 & 0.62 &    - &    - &    - &    - \\
& BoN (N-RM)        & 33.4 \textcolor{ImproveGreen}{↑67.0\%} & 32.0 & 1.75 & 0.29 & 25.0 \textcolor{ImproveGreen}{↑56.3\%} & 32.0 & 1.76 & 0.80 \\
& BoN (S-RM)        & 31.0 \textcolor{ImproveGreen}{↑55.0\%} & 32.0 & 1.72 & 0.25 & 24.0 \textcolor{ImproveGreen}{↑50.0\%} & 32.0 & 1.65 & 0.55 \\
& \cellcolor{lightgray!30}\textbf{BEACON (N-RM)} & \cellcolor{lightgray!30}32.8 \textcolor{ImproveGreen}{↑64.0\%} & \cellcolor{lightgray!30}15.8 \textcolor{ReduceBlue}{↓50.6\%} & \cellcolor{lightgray!30}1.68 & \cellcolor{lightgray!30}\textbf{1.12} & \cellcolor{lightgray!30}23.5 \textcolor{ImproveGreen}{↑46.9\%} & \cellcolor{lightgray!30}14.5 \textcolor{ReduceBlue}{↓54.7\%} & \cellcolor{lightgray!30}1.68 & \cellcolor{lightgray!30}\textbf{1.20} \\
& \cellcolor{lightgray!30}\textbf{BEACON (S-RM)} & \cellcolor{lightgray!30}32.0 \textcolor{ImproveGreen}{↑60.0\%} & \cellcolor{lightgray!30}16.1 \textcolor{ReduceBlue}{↓49.7\%} & \cellcolor{lightgray!30}1.66 & \cellcolor{lightgray!30}\textbf{1.08} & \cellcolor{lightgray!30}22.5 \textcolor{ImproveGreen}{↑40.6\%} & \cellcolor{lightgray!30}14.8 \textcolor{ReduceBlue}{↓53.8\%} & \cellcolor{lightgray!30}1.53 & \cellcolor{lightgray!30}\textbf{1.15} \\
\midrule
\multirow{8}{*}{\texttt{Qwen2.5-7B}}
& Direct CoT     & 43.0 &  1.0 & -1.25 & -0.53 & 22.5 &  1.0 & -0.85 & -0.53 \\
& SC                & 50.0 \textcolor{ImproveGreen}{↑16.3\%} & 16.0 \textcolor{ReduceBlue}{↓50.0\%} & -0.37 & -0.02 &    - &    - &    - &    - \\
& RASC              & 49.5 \textcolor{ImproveGreen}{↑15.1\%} &  4.3 \textcolor{ReduceBlue}{↓86.6\%} & -0.28 & 0.57 & 25.0 \textcolor{ImproveGreen}{↑11.1\%} &  4.0 \textcolor{ReduceBlue}{↓87.5\%} & -1.15 & -0.30 \\
& AS                & 49.3 \textcolor{ImproveGreen}{↑14.7\%} &  4.6 \textcolor{ReduceBlue}{↓85.6\%} & -0.29 & 0.55 &    - &    - &    - &    - \\
& BoN (N-RM)        & 55.2 \textcolor{ImproveGreen}{↑28.4\%} & 32.0 & 1.85 & 0.09 & 36.0 \textcolor{ImproveGreen}{↑60.0\%} & 32.0 & 2.02 & 1.02 \\
& BoN (S-RM)        & 55.0 \textcolor{ImproveGreen}{↑27.9\%} & 32.0 & 1.76 & 0.05 & 35.0 \textcolor{ImproveGreen}{↑55.6\%} & 32.0 & 2.05 & 1.05 \\
& \cellcolor{lightgray!30}\textbf{BEACON (N-RM)} & \cellcolor{lightgray!30}54.0 \textcolor{ImproveGreen}{↑25.6\%} & \cellcolor{lightgray!30}6.5 \textcolor{ReduceBlue}{↓79.7\%} & \cellcolor{lightgray!30}1.78 & \cellcolor{lightgray!30}\textbf{0.92} & \cellcolor{lightgray!30}33.5 \textcolor{ImproveGreen}{↑48.9\%} & \cellcolor{lightgray!30}7.8 \textcolor{ReduceBlue}{↓75.6\%} & \cellcolor{lightgray!30}1.95 & \cellcolor{lightgray!30}\textbf{1.55} \\
& \cellcolor{lightgray!30}\textbf{BEACON (S-RM)} & \cellcolor{lightgray!30}54.0 \textcolor{ImproveGreen}{↑25.6\%} & \cellcolor{lightgray!30}7.0 \textcolor{ReduceBlue}{↓78.1\%} & \cellcolor{lightgray!30}1.64 & \cellcolor{lightgray!30}\textbf{0.91} & \cellcolor{lightgray!30}33.0 \textcolor{ImproveGreen}{↑46.7\%} & \cellcolor{lightgray!30}8.0 \textcolor{ReduceBlue}{↓75.0\%} & \cellcolor{lightgray!30}1.90 & \cellcolor{lightgray!30}\textbf{1.50} \\
\midrule
\multirow{8}{*}{\texttt{Grok-3-Mini}}
& Direct CoT     & 89.0 &  1.0 & -1.10 & -0.28 & 82.0 &  1.0 & -0.98 & -0.70 \\
& SC                & 92.0 \textcolor{ImproveGreen}{↑3.4\%} & 16.0 \textcolor{ReduceBlue}{↓50.0\%} & -0.38 & -0.04 &    - &    - &    - &    - \\
& RASC              & 93.8 \textcolor{ImproveGreen}{↑5.4\%} &  3.5 \textcolor{ReduceBlue}{↓89.1\%} & -0.22 & 0.56 & 85.5 \textcolor{ImproveGreen}{↑4.3\%} &  3.5 \textcolor{ReduceBlue}{↓89.1\%} & -1.02 & -0.55 \\
& AS                & 93.8 \textcolor{ImproveGreen}{↑5.4\%} &  3.8 \textcolor{ReduceBlue}{↓88.1\%} & -0.23 & 0.53 &    - &    - &    - &    - \\
& BoN (N-RM)        & 95.5 \textcolor{ImproveGreen}{↑7.3\%} & 32.0 & 1.62 & 0.10 & 94.0 \textcolor{ImproveGreen}{↑14.6\%} & 32.0 & 1.45 & 0.80 \\
& BoN (S-RM)        & 95.0 \textcolor{ImproveGreen}{↑6.7\%} & 32.0 & 1.70 & 0.19 & 94.2 \textcolor{ImproveGreen}{↑14.9\%} & 32.0 & 1.42 & 0.79 \\
& \cellcolor{lightgray!30}\textbf{BEACON (N-RM)} & \cellcolor{lightgray!30}94.8 \textcolor{ImproveGreen}{↑6.5\%} & \cellcolor{lightgray!30}5.0 \textcolor{ReduceBlue}{↓84.4\%} & \cellcolor{lightgray!30}1.57 & \cellcolor{lightgray!30}\textbf{0.99} & \cellcolor{lightgray!30}92.8 \textcolor{ImproveGreen}{↑13.2\%} & \cellcolor{lightgray!30}4.0 \textcolor{ReduceBlue}{↓87.5\%} & \cellcolor{lightgray!30}1.39 & \cellcolor{lightgray!30}\textbf{1.94} \\
& \cellcolor{lightgray!30}\textbf{BEACON (S-RM)} & \cellcolor{lightgray!30}94.2 \textcolor{ImproveGreen}{↑5.8\%} & \cellcolor{lightgray!30}5.5 \textcolor{ReduceBlue}{↓82.8\%} & \cellcolor{lightgray!30}1.64 & \cellcolor{lightgray!30}\textbf{0.95} & \cellcolor{lightgray!30}93.4 \textcolor{ImproveGreen}{↑13.9\%} & \cellcolor{lightgray!30}4.3 \textcolor{ReduceBlue}{↓86.6\%} & \cellcolor{lightgray!30}1.36 & \cellcolor{lightgray!30}\textbf{1.93} \\
\bottomrule
\multicolumn{10}{@{}l}{\textit{Note:} SC = Self-Consistency \citep{wang2022self}; 
RASC = Reasoning-Aware Self-Consistency \citep{wan-etal-2025-reasoning};} \\
\multicolumn{10}{@{}l}{AS = Adaptive Sampling \citep{aggarwal-etal-2023-lets}; 
BoN = Best-of-$N$ sampling \citep{cobbe2021training} SC and AS are not applicable to alignment tasks.} \\
\end{tabular}
\end{table*}


\subsection{Main Results}
\label{sec:results_inference}
\textbf{Setup.} We use a warm-start of $k_0=3$ (to initialize a Jeffreys' prior) and maximum horizon of $n=32$ (standard for Best-of-$N$ \citep{singhi2025solveverifycomputeoptimalproblem}). For each method $m$, $K_m$ denotes its realized sample count ($1 \le K_m \le n$) and $\overline{K}_m$ its dataset average. We focus comparisons on training-free methods and 
standard BoN as these represent the most widely deployed approaches. Specifically, we include one-shot \textbf{Chain-of-Thought} (CoT) with $K{=}1$; \textbf{Self-Consistency} (SC) \citep{wang2022self} with majority vote over $n$ samples; \textbf{Adaptive-Consistency} (AS) \citep{aggarwal-etal-2023-lets}, a model-agnostic method that fits a Dirichlet–multinomial posterior to agreement patterns for early stopping; \textbf{RASC} \citep{wan-etal-2025-reasoning}, a heuristic adaptive sampler using CoT quality scores; and Best-of-$N$ (\textbf{BoN}), which selects the highest reward-scored candidate from $n$ attempts. Policy models include LLaMA-3.2-8B, Qwen2.5-7B-Instruct, and Grok-3-mini; reward models include Llama-3.1-Nemotron-70B-Reward and Skywork-Llama-3.1-8B. We used CoT \citep{wei2022chain} prompting with model-specific templates. Evaluation covered: (1) Pass @ 1 for \textit{Reasoning Tasks} on three mathematical benchmarks (\texttt{MATH-500} \citep{lightman2023let}, \texttt{AIME24} \citep{AIME2024}, \texttt{AMC23}); (2) \textit{Alignment Task} using \texttt{AlpacaEval 2.0} \citep{alpaca_eval}, comparing responses for user instructions against GPT-4; and (3) for both tasks, \textit{expected standardized reward} $\bar{\hat{z}}_K$ and \textit{expected standardized value} $\bar{\hat{V}}_K = \mathbb{E}[z_K - K\cdot c]$, representing the quality-efficiency tradeoff (more details in Appendix~\ref{app:main_setup}).

\noindent \textbf{Optimal Performance-Efficiency Tradeoff.}
Table~\ref{tab:model_comparison_no_siunitx_updated} demonstrates BEACON's ability to achieve an optimal balance between performance and computational efficiency. Our approach consistently matches BoN's performance while requiring significantly fewer samples. This tradeoff is quantitatively validated by \emph{consistently superior value function scores}, confirming that BEACON effectively maximizes expected net gain as a Bayesian optimal stopping solution. Results remain consistent across experimental conditions—including different base models, reward models, and task categories.

\begin{figure}[t]
  \centering
  \includegraphics[width=0.91\linewidth]{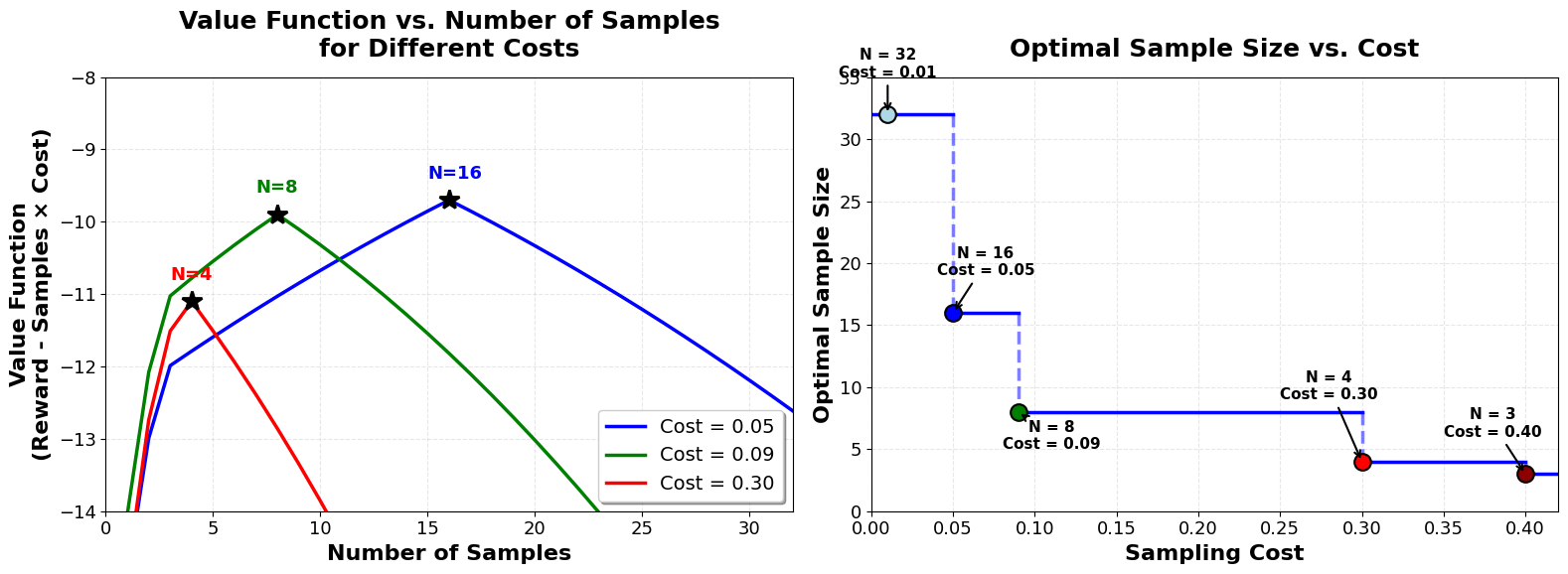} 
  \captionof{figure}{Impact of the sampling cost ($c$) on the value optimization and the optimal sample size. Higher $c$ results in earlier stopping to reach Bayesian optimality.}
  \label{fig:cost_analysis_value_functions}
\end{figure}

\noindent \textbf{Impact of Sampling Cost ($c$) on Optimized Sample Size.}
We analyze how the sampling cost parameter $c$ shapes BEACON’s adaptive sample size $K_{BEACON}$. As shown in Figure~\ref{fig:cost_analysis_value_functions} , increasing $c$ consistently reduces the optimal number of samples, aligning with our discussion in Section \ref{sec:model}. Unlike factors such as reward variance, response quality, or remaining budget—which emerge from query characteristics—$c$ serves as a \textit{human-interpretable} control knob for balancing efficiency and quality. Stopping typically occurs when responses stabilize, when quality remains uniformly low, or when nearing the maximum budget (examples in ~\ref{app:diversity}). For practical deployment, we recommend a default $c=0.1$ when there is no strong preference between efficiency and quality. Lower values of $c$ are better suited for difficult, high-variance tasks, whereas higher values suit easier or consistent ones (See App.~\ref{app:hyperparameter_analysis} for guidelines).

\begin{figure}[t]
  \centering
  \includegraphics[width=0.88\linewidth]{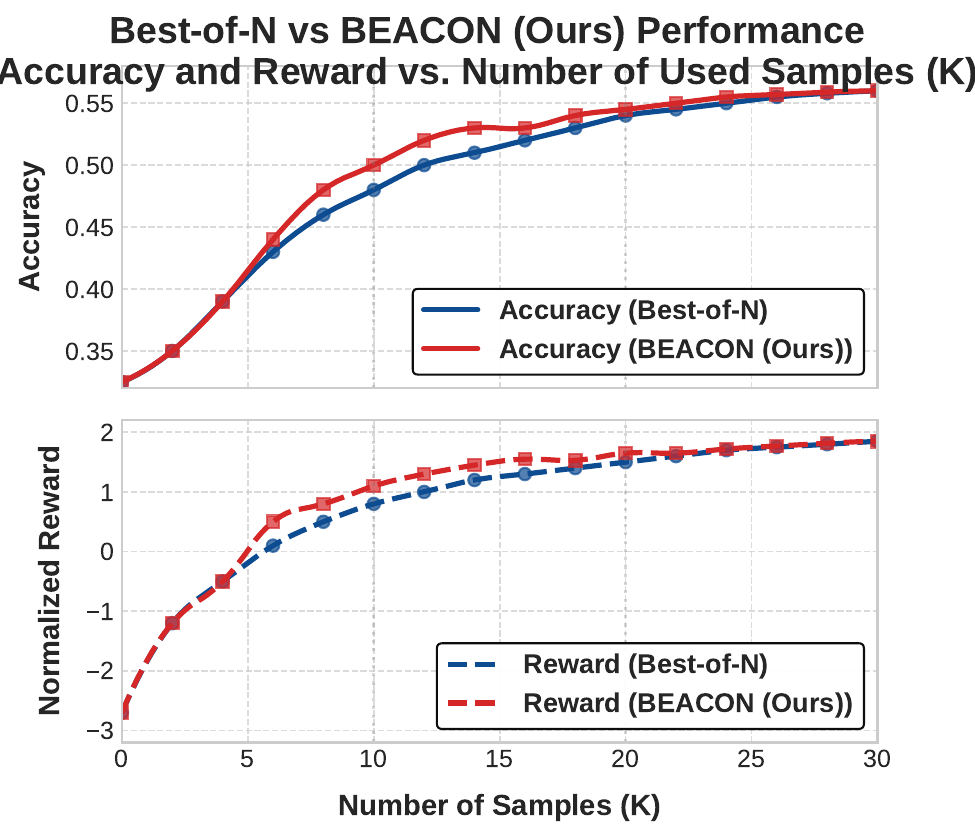}
  \caption{\textbf{Comparative performance trajectories on Best-of-N vs.~BEACON}. When forcing the same number of samples, BEACON achieves better performance; When restricting on a threshold performance, BEACON would require less number of samples.}
  \vspace{-8pt}
  \label{fig:bon_vs_beacon_trajectories_controlled}
\end{figure}

\noindent \textbf{BEACON's Effectiveness at Controlled Average Sample Sizes.}
\label{sec:beacon_controlled_K}
Building on our analysis of how sampling cost $c$ influences BEACON's stopping behavior, we provide an alternative perspective by investigating scenarios where $c$ is calibrated to ensure BEACON's average sample size ($\overline{K}_{BEACON}$) matches the fixed sample size of standard Best-of-N (BoN) strategies. Figure~\ref{fig:bon_vs_beacon_trajectories_controlled} illustrates this controlled comparison, revealing BEACON's advantages: (1) when using the \textit{same number of samples} ($\overline{K}_{BEACON} = K_{BoN}$), BEACON achieves substantially \textit{higher accuracy and reward}; and (2) BEACON can maintain \textit{equivalent accuracy and reward} while requiring \textit{fewer samples}. This performance advantage stems from BEACON's dynamic sampling strategy, which intelligently invests additional samples in promising queries while terminating earlier for less promising ones—a fundamental improvement over BoN's uniform sampling approach that allocates identical resources regardless of query nature or potential quality improvements.

\subsection{Extensions and Applications}\label{sec:extensions}

\textbf{Efficiency in Data Generation for Iterative DPO.}
\label{exp:idpo}To demonstrate BEACON's practical utility, we applied it to improve efficiency in data generation for Iterative Direct Preference Optimization (DPO) \citep{Rafailov2023DirectPO}, which enhances LLM consistency on challenging questions \citep{xiong2025building}. We evaluated BEACON against a standard Best-of-N (BoN) approach—the conventional method for generating preference pairs in iterative training processes \citep{yuan2025selfrewardinglanguagemodels}; comprehensive experimental details are provided in Appendix~\ref{app:dpo_implementation_details}. Figure~\ref{fig:dpo_comparison} illustrates performance across seven DPO iterations, revealing that BEACON achieved \emph{comparable accuracy} while progressively reducing the average required samples ($\overline{K}$, shown on the secondary y-axis). This preliminary experiment demonstrates that as LLMs become more consistent and aligned with preferences, BEACON efficiently identifies high-quality preference pairs with substantially fewer samples to improve post-training efficiency.

\begin{figure}[t]
  \centering
  \includegraphics[width=0.88\linewidth]{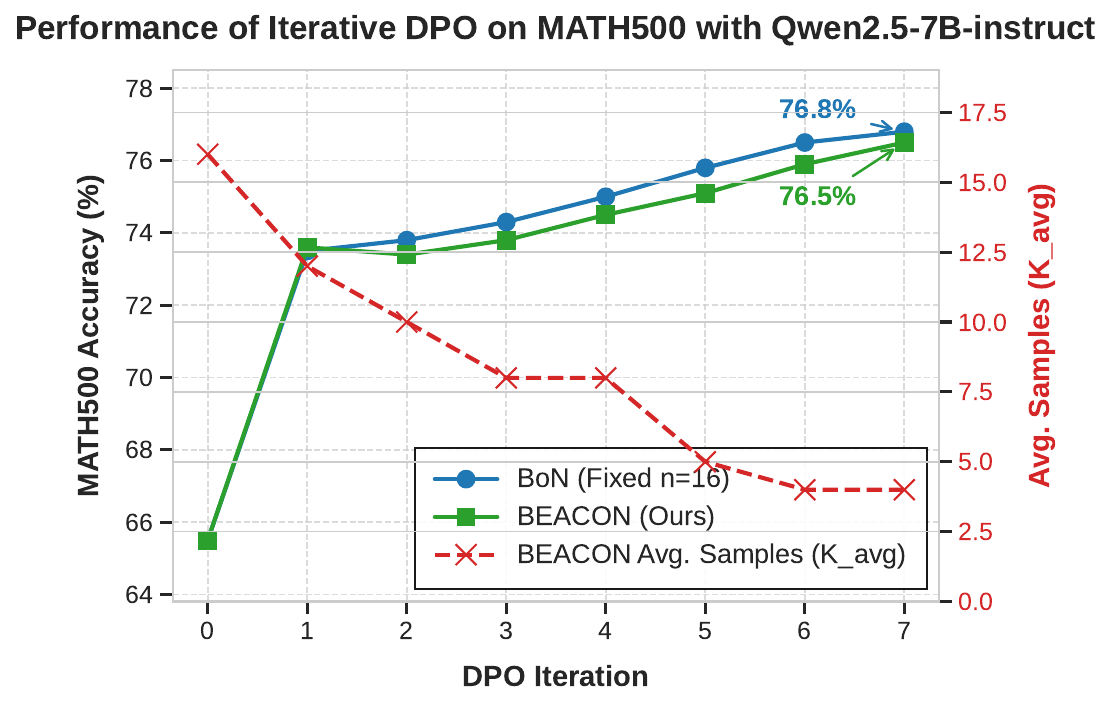}
  \caption{Iterative DPO performance on MATH500 with Qwen2.5-7B-instruct. BEACON (accuracy: green squares) achieves competitive accuracy relative to BoN (fixed $n{=}16$, accuracy: blue circles) across seven DPO iterations while reducing the average number of samples required per prompt ($\overline{K}$: red dashed line, right y-axis).}
  \vspace{-8pt}
  \label{fig:dpo_comparison}
  
\end{figure}

\noindent \textbf{Robust Updating for Negative Skewness.}  Reward distributions from LLMs occasionally exhibit negative skewness (Fig \ref{fig:normality_corr}), where extreme low outliers distort posterior updates. We introduce a \emph{robust updating mechanism} that preserves the informative right tail—critical for identifying maximum rewards—while mitigating left-tail outliers. Values below the 1\% posterior-predictive quantile are replaced with the current posterior mean, ensuring high-quality candidates remain intact while the posterior better reflects the reward landscape. This adaptive update reduces outlier impact without violating Gaussian assumptions. Figure~\ref{fig:side_by_side_adaptive_plot} shows the robust mechanism consistently drives BEACON's stopping points closer to the true optimum. Full derivations and analysis are in Appendix~\ref{app:robust_update}.

\noindent \textbf{Parallel Sampling.} \label{exp:batch} While BEACON is inherently sequential, practical deployments can benefit from batch-parallel sampling, like BoN Sampling, to reduce wall-clock time at the expense of higher memory usage. In this mode, BEACON generates batches of $b$ samples simultaneously, \emph{updates posterior beliefs after each batch}, and applies the same stopping criterion $h_{n,k}(\hat{z}_k) \leq \tfrac{c}{\sigma_k}$, where $k$ now indexes completed batches. Within each batch, responses are sampled independently using the same prompt, then evaluated with the reward model; posterior parameters are updated with all batch rewards, and the stopping rule is applied. Termination selects the highest-reward response across all batches. As shown in Table~\ref{tab:parallel_performance}, batching can slightly improve accuracy since larger batches enforce a minimum exploration depth before each stopping decision, though this comes with diminishing returns and increased memory overhead.  Moreover, as discussed in \ref{met:comp}, BEACON’s pre-computed UIP index tables can be \emph{reused across multiple parallel queries} without recomputation, enabling efficient query-level parallelization at no additional cost.

\begin{table}[t]
\centering
\small
\vspace{3pt}
\caption{Batch-parallel BEACON performance across different batch sizes, evaluated with LLaMA~3.2 on a mathematical reasoning dataset. Memory overhead is measured relative to sequential execution ($b=1$).}
\label{tab:parallel_performance}
\begin{tabular}{@{}lcccc@{}}
\toprule
\textbf{Batch} & \textbf{Speedup} & \textbf{Memory} & \textbf{Acc.} & \textbf{Avg.} \\
\textbf{Size} & (vs Seq.) & Overhead & Change & \textbf{Samples} \\
\midrule
1 (Seq.) & 1.00$\times$ & Baseline & 0.0\% & 15.8 \\
2 & 1.65$\times$ & +12\% & +0.2\% & 16.1 \\
4 & 2.31$\times$ & +28\% & +0.6\% & 17.2 \\
8 & 2.85$\times$ & +65\% & +0.8\% & 19.4 \\
\bottomrule
\end{tabular}
\vspace{-7pt}
\end{table}

\section{Related Work}

\textbf{Parallel Reasoning and Efficiency.} Parallel scaling approaches \citep{zeng2025revisitingtesttimescalingo1like,qian2025scaling} enhance LLM answers by generating multiple candidates and aggregating them into a final answer, through consensus (e.g., majority voting and weighted confidence scores \citep{wang2022self,chen2024universal,fu2025deep}) or with external verifiers and reward models that rank and select superior solutions \citep{cobbe2021training,ichihara2025evaluation,zhang2024generative,ankner2024critiqueoutloud}. While effective, these methods prioritize response quality without explicitly addressing computational costs. Recent work on efficient sampling strategies \citep{Sui2025StopOverthinking,Fu2024Certaindex} seeks to address the issue using fine-tuned verifiers \citep{manvi2024adaptiveinferencetimecomputellms,huang2025efficienttesttimescalingselfcalibration,wan-etal-2025-reasoning} or heuristic rules that adapt resource allocation by query complexity \citep{wang2025samplingefficienttesttimescalingselfestimating,wan2025derailerreraileradaptiveverificationefficient,aggarwal-etal-2023-lets}. In contrast, our BEACON framework takes a principled Bayesian learning approach that integrates reward signals with optimal stopping theory to jointly optimize response quality and computational efficiency.

\noindent \textbf{Bandits and Bayesian Optimization.}
BEACON’s sequential search framework connects to established paradigms in decision theory and optimization \citep{keith2021survey}. While Multi-Armed Bandit problems \citep{lattimore2020bandit,slivkins2019introduction} emphasize maximizing cumulative rewards, BEACON focuses on finding the maximum reward with minimal sampling. Our work builds more directly on best-arm identification \citep{audibert2010best,gabillon2012best} and Extreme Bandits \citep{carpentier2014extreme,lopez2021learning}, which similarly target optimal or extreme-value outcomes. BEACON’s novelty lies in combining a Bayesian approach with conjugate priors for adaptive belief updates and optimal stopping theory \citep{ferguson2012optimal} to decide when further exploration is no longer cost-effective. While conceptually related to budgeted bandits \citep{xia2016budgeted} and Bayesian optimization \citep{shahriari2015taking}, BEACON  applies these ideas to LLM sample efficiency, bridging theoretical insights with practical inference.

\section{Conclusion}
\label{sec:conclusion}
We introduced \textbf{BEACON}, a principled framework grounded in sequential search theory with Bayesian learning under conjugate priors. BEACON addresses the fundamental trade-off between computational cost and response quality during LLM inference by dynamically determining when to stop sampling based on evolving posterior beliefs about reward distributions and the cost of additional sampling. With strong empirical results and comprehensive theoretical analysis, we demonstrate the value of decision-theoretic approaches for resource-aware scaling in LLM reasoning and generation.

\newpage

\section*{Limitations}
\label{sec:limitations}

While BEACON provides strong efficiency gains, several avenues remain 
for extending the framework. First, BEACON's optimality guarantees assume 
the reward model provides accurate quality signals; systematic reward model 
miscalibration could lead to suboptimal stopping decisions. Integrating 
reward model uncertainty quantification—for instance, by incorporating 
ensemble-based confidence estimates into the posterior updates—represents 
a promising direction for enhancing robustness. Second, the framework 
currently assumes independence across samples, whereas exploiting 
correlations between generated responses (e.g., through shared reasoning 
patterns) could further improve sample efficiency.

Future research can address these opportunities by extending BEACON to 
leverage reward model ensembles for uncertainty-aware stopping and 
incorporating structured priors that capture dependencies between samples. 
Furthermore, our preliminary experiments with Iterative DPO (Section~\ref{exp:idpo}) 
suggest BEACON's potential for post-training optimization, warranting 
deeper investigation into its role across different training paradigms and 
reward model architectures. Additionally, dynamic tuning of the cost 
parameter $c$ based on query characteristics could enable fully automated 
adaptation to varying computational budgets. Together, these directions 
position BEACON as a foundation for both efficient inference and adaptive 
post-training pipelines in production deployments.  


\bibliography{custom}

\appendix

\section{Experimental Setup and Details}
\label{app:experimental_setup_details}

\subsection{Main Experiment Setup}
\label{app:main_setup}

\subsubsection{Policy Models and Reward Models}
\label{app:models_used}

For our policy (generator) models, we evaluated a range of architectures and sizes to ensure comprehensive assessment of BEACON's performance:
\begin{itemize}
    \item \textbf{LLaMA-3.2-3B-Instruct}: Accessed and run locally on a GPU node equipped with 2 NVIDIA A100 GPUs.
    \item \textbf{Qwen2.5-7B-Instruct}: Inference conducted via DeepInfra's API\footnote{\url{https://deepinfra.com/}}.
    \item \textbf{Grok-3-Mini}: Inference conducted via Grok's API\footnote{\url{https://grok.x.ai/}}.
\end{itemize}

For our reward models (RMs), we utilized:
\begin{itemize}
    \item \textbf{NVIDIA Llama-3.1-Nemotron-70B-Reward}: Accessed via NVIDIA's API \footnote{\url{https://build.nvidia.com/nvidia/llama-3_1-nemotron-70b-reward}} services for response verification.
    \item \textbf{Skywork-Llama-3.1-8B-Reward}: Accessed and run locally on a GPU node equipped with 2 NVIDIA A100 GPUs.
\end{itemize}

\subsubsection{Datasets and Evaluation Benchmarks}
\label{app:datasets_benchmarks}

We evaluated BEACON on diverse reasoning and alignment tasks:

\paragraph{Reasoning Tasks:}
The reasoning evaluation focused on mathematical problem-solving, reporting average accuracy (Pass@1) across the following benchmarks. For answer extraction and checking mathematical equivalence, we utilized the expression matching tool from the Math-Verify repository\footnote{\url{https://github.com/huggingface/Math-Verify}} to ensure consistent and robust evaluation.
\begin{itemize}
    \item \textbf{MATH500}: We used a randomly selected subset of 50 problems from the MATH500 dataset~\citep{hendrycksmath2021} to provide a broad assessment without over-focusing on this specific dataset, given its large size. The problems are sampled from the test set.
    \item \textbf{AIME 2024}: All 30 problems from the American Invitational Mathematics Examination 2024 were used~\citep{AIME2024}..
    \item \textbf{AMC 2023}: All 40 problems from the American Mathematics Competitions 2023 were used, combining problems from AMC 10 and AMC 12.
\end{itemize}

\paragraph{Alignment Task:}
\begin{itemize}
    \item \textbf{AlpacaEval 2.0}: We used the full set of 805 prompts from AlpacaEval 2.0~\citep{alpaca_eval}. The evaluation of response quality, comparing model outputs against a reference (e.g., GPT-4), was conducted using OpenAI's API for the automated evaluation protocol provided by AlpacaEval. The primary metric reported is the win rate.
\end{itemize}

\subsubsection{Prompting Strategy}
\label{app:prompting_strategy}
For all reasoning tasks, we employed a standard one-shot Chain-of-Thought (CoT) prompting strategy. The models were instructed to act as math assistants and provide step-by-step solutions. Model-specific instruction templates were used where appropriate, but the core reasoning prompt structure was as follows:

\begin{tcolorbox}[colback=gray!5, colframe=black!75, title=Standard CoT Reasoning Prompt]
You are a math assistant. Solve problems step by step with clear reasoning.

\textbf{Format:}
\begin{enumerate}
\item Start with ``Let me solve this step by step:''
\item Numbered steps with explanations
\item End with answer in box: $\boxed{42}$
\end{enumerate}

\textbf{Rules:}
\begin{itemize}
\item Answer must be integer or simplified fraction
\item Use exact box format: $\boxed{42}$
\item No text after box
\item For fractions: $\boxed{\frac{3}{4}}$
\item For negatives: $\boxed{-5}$
\end{itemize}

\textbf{Example:}

Let me solve this step by step:
\begin{enumerate}
\item First step
\item Second step
\item[\vdots]
\item[N)] Final step
\end{enumerate}

$$\boxed{42}$$

\textbf{[Problem Statement Here]}
\end{tcolorbox}

This standardized prompt ensures consistency in how tasks are presented to the policy models. For AlpacaEval, prompts are used as provided by the benchmark.

\section{Practical Guidelines for Setting the hyperamareters}
\label{app:cost_guide}

The sampling cost parameter $c$ plays a central role in BEACON, and should be viewed as the user's tolerance towards both time and computational resources required for an additional sampling. As demonstrated in our main results, different values of $c$ directly shape the optimization landscape of the value function $E[z_K - K \cdot c]$, with higher $c$ values consistently leading to earlier stopping on average. Based on our comprehensive experiments across multiple models and tasks, we recommend setting $c = 0.1$ as an effective starting point, as this value achieves significant sample reduction (approximately 50-80\% fewer samples than fixed BoN) while preserving comparable accuracy for both reward models across different task categories. (highlight h-index table) to sample them in parallel without constraint.

To determine the optimal $c$ for specific deployment requirements, we recommend the following calibration procedure:

\begin{enumerate}
    \item \textbf{Establish performance baselines:} First, run a small-scale experiment (e.g., 50-100 queries) using fixed BoN with a large sample size (e.g., $N=32$) to establish upper-bound performance metrics (accuracy, reward).
    
    \item \textbf{Sweep across $c$ values:} Conduct a parameter sweep across a range of $c$ values (e.g., $c \in \{0.01, 0.05, 0.1, 0.2, 0.3, 0.4\}$) using BEACON on the same query set.
    
    \item \textbf{Analyze the performance-efficiency Pareto frontier:} For each $c$ value, plot the resulting performance metric (e.g., accuracy) versus average sample count $\bar{K}$. The optimal $c$ lies at the "knee point" of this curve where marginal performance gains begin to diminish relative to increased sampling costs.
\end{enumerate}

In our experiments, we observed distinct patterns across different model sizes and tasks:

\begin{itemize}
    \item For smaller models (e.g., LLaMA-3.2-3B) on reasoning tasks, $c \approx 0.1$ typically reduced samples by $\sim$50\% while maintaining accuracy within 1-2\% of the full BoN baseline.
    
    \item For larger models (e.g., Grok-3-Mini) on alignment tasks, $c \approx 0.3$ was often sufficient, reducing samples by $\sim$85\% with minimal performance degradation, as these models generally produced more consistent high-quality responses. Note that to ensure fair comparison we still adapt same cost $c = 0.1$ for these models.
    
    \item For time-sensitive applications (e.g., interactive chatbots), higher values ($c \approx 0.2$-$0.3$) prioritize responsiveness.
    
    \item For high-stakes applications (e.g., critical reasoning tasks), lower values ($c \approx 0.05$) favor thoroughness.
\end{itemize}

The optimal value of $c$ is inherently subjective, as some users may have stricter resource constraints or lower tolerance for computation time, while others may prioritize response quality over efficiency. In production environments, $c$ can be further contextualized to correspond to actual computational costs.

Again we want to highlight that the key advantage of BEACON is that regardless of how $c$ is set, our framework mathematically guarantees that no resources are wasted through over-sampling or under-sampling, given the specific resource constraint expressed through $c$. This adaptivity ensures BEACON consistently delivers the optimal performance-efficiency trade-off aligned with the user's particular tolerance for computational cost. Furthermore, $c$ can be dynamically adjusted based on changing conditions, such as server load, time of day, or query importance.

\subsection{Hyperparameter Selection Guide}
\label{app:hyperparameter_analysis}

Based on the above analysis and guidandec, we provide Table~\ref{tab:hyperparameter_sensitivity} which provides systematic guidance for selecting optimal hyperparameters across different application scenarios and model configurations.

\begin{table*}[htbp]
\centering
\caption{Hyperparameter Selection Guidelines}
\label{tab:hyperparameter_sensitivity}
\begin{tabular}{llccccc}
\toprule
\textbf{Application} & \textbf{Model Size} & \textbf{Optimal} & \textbf{Horizon} & \textbf{Expected} & \textbf{Accuracy} & \textbf{Primary} \\
\textbf{Type} & \textbf{(Params)} & \textbf{Cost $(c)$} & \textbf{$(n)$} & \textbf{Samples $(K)$} & \textbf{(\%)} & \textbf{Objective} \\
\midrule
Easy Reasoning & $\leq$7B & 0.15 & 16 & 4.2 & 85.1 & Efficiency \\
Easy Reasoning & $\geq$70B & 0.30 & 16 & 3.1 & 87.4 & Speed \\
Hard Reasoning & $\leq$7B & 0.05 & 32 & 12.8 & 52.9 & Quality \\
Hard Reasoning & $\geq$70B & 0.10 & 32 & 8.4 & 58.2 & Balanced \\
Creative Writing & $\leq$7B & 0.08 & 24 & 9.6 & 76.3 & Diversity \\
Creative Writing & $\geq$70B & 0.20 & 16 & 5.2 & 82.1 & Consistency \\
Code Generation & $\leq$7B & 0.06 & 32 & 14.1 & 68.7 & Correctness \\
Code Generation & $\geq$70B & 0.12 & 24 & 7.8 & 75.4 & Efficiency \\
Interactive Chat & Any & 0.40 & 12 & 3.5 & 71.2 & Latency \\
Batch Processing & Any & 0.05 & 32 & 18.2 & 55.8 & Thoroughness \\
\bottomrule
\end{tabular}
\end{table*}

\subsection{Cost Parameter Calibration}
\label{app:cost_calibration}

Table~\ref{tab:cost_calibration} demonstrates the systematic relationship between cost parameter values and resulting performance metrics, enabling data-driven hyperparameter selection.

\begin{table*}[htbp]
\centering
\caption{Cost Parameter Calibration Analysis}
\label{tab:cost_calibration}
\begin{tabular}{cccccc}
\toprule
\textbf{Cost $(c)$} & \textbf{Avg. Accuracy} & \textbf{Avg. Samples} & \textbf{Value Score} & \textbf{95\% Sample} & \textbf{Recommended} \\
 & \textbf{(\%)} & \textbf{$(K)$} & \textbf{$E[z_K - K \cdot c]$} & \textbf{Count} & \textbf{Scenario} \\
\midrule
0.01 & 54.2 & 28.4 & 0.94 & 32.0 & Research/Quality-critical \\
0.05 & 53.4 & 18.6 & 1.41 & 28.5 & Hard reasoning tasks \\
0.10 & 52.9 & 12.8 & 1.61 & 22.1 & \textbf{Default (balanced)} \\
0.20 & 51.8 & 8.2 & 1.54 & 16.4 & General applications \\
0.30 & 50.1 & 5.9 & 1.33 & 12.8 & Latency-sensitive \\
0.50 & 47.8 & 4.1 & 1.02 & 8.9 & Interactive systems \\
\bottomrule
\end{tabular}
\end{table*}

In practice, $c$ can be calibrated to actual costs: for API-based 
inference, set $c$ proportional to (\$/token) $\times$ (avg.\ tokens per sample); 
for self-hosted deployment, $c \propto$ (GPU hours) $\times$ (hourly cost). 
For instance, if generating one sample costs \$0.01 and the reward scale 
is approximately $[-3, 3]$, setting $c = 0.1$ implies stopping when the 
marginal expected gain falls below \$0.01.

\section{Practical Implementation Details}
\label{app:practical_extensions}

\subsection{Decoding Hyperparameters and BEACON Configuration}
\label{app:hyperparameters_beacon_config}

Besides cost, the BEACON framework operated with the following core settings for the main experiments:

\paragraph{Decoding Parameters:} For all policy model inferences, unless specified otherwise by the API provider's defaults for instructed models, we used consistent decoding parameters:
\begin{itemize}
    \item Temperature: $0.7$
    \item Maximum new tokens: $2048$
\end{itemize}

\paragraph{BEACON Framework Configuration:}

\begin{itemize}
    \item Minimum initial samples ($k_0$): $3$. This is required to properly establish the posterior Normal-Inverse-Gamma distribution using Jeffreys' non-informative prior ($\mu_0=0, \nu_0=0, \alpha_0=-0.5, \beta_0=0$).
    \item Maximum sampling horizon ($T_{\max}$ or $n$): $32$. 
    \item H-index function $h_{n,k}(\hat{z}_k)$: Pre-computed based on the methodology in Baucells and Zorc~\citep{baucells2024search} using the specified priors.
\end{itemize}

\subsection{Iterative DPO with BEACON: Experimental Setup}
\label{app:dpo_implementation_details} 

\subsubsection{Data Foundation for DPO Preference Generation}
\label{app:data_foundation_dpo}
The prompts used for generating responses, which subsequently form preference pairs for Direct Preference Optimization (DPO), are drawn from the extensive training set of the MATH dataset (7,500 problems) \citep{hendrycksmath2021}. The MATH dataset, comprising problems from American mathematics competitions like AMC 10, AMC 12, and AIME, is highly suitable due to its provision of full step-by-step solutions, which is beneficial for training models on complex derivations. Its effectiveness in enhancing mathematical reasoning is well-established. While preference data is generated from these MATH prompts, the DPO model's performance is evaluated on the MATH500 benchmark, as presented in Figure~\ref{fig:dpo_comparison} of the main text.

\subsubsection{Fixed Reward Model for Ranking Responses}
\label{app:fixed_reward_model}
To rank the generated responses for constructing preference pairs $(y_w, y_l)$ needed for DPO, simiarly to the main experiment, we use Skywork-llama3.1-8b as the base reward model. This reward model assigns a scalar score, denoted $R_{learned}$, to each policy-generated response, reflecting its assessed quality.

\subsubsection{Preference Pair Generation Strategies for DPO}
\label{app:preference_generation_dpo}
Preference pairs for each DPO iteration are generated using one of two distinct strategies:

\paragraph{Best-of-N (BoN) Strategy.}
For each prompt, a fixed $N=16$ candidate responses are sampled from the current iteration of the policy model (Qwen2.5-7B-instruct). These $N$ responses are then scored using the $R_{learned}$ value obtained from our reward model. The response with the highest $R_{learned}$ score is selected as the preferred response ($y_w$), and the response with the lowest $R_{learned}$ score is chosen as the dispreferred response ($y_l$). If all 16 responses for a given prompt yield identical $R_{learned}$ scores, that prompt is excluded from the DPO training set for that iteration.

\paragraph{BEACON Strategy.}
For each prompt, our BEACON algorithm is employed to adaptively determine the number of samples $K$ to generate. BEACON begins with an initial $k_0=3$ samples and can sample up to a maximum of $N_{max}=16$ (especially in early DPO iterations). As noted in the main text (Figure~\ref{fig:dpo_comparison}, right panel), the average $K$ required by BEACON tends to decrease in later DPO iterations. BEACON utilizes an adapted reward signal, $R_{BEACON}$ (the transformation from $R_{learned}$ is detailed in Appendix~\ref{sssec:reward_adaptation_beacon}), for both its internal Bayesian stopping decisions and for the final ranking of the $K$ collected samples. From these $K$ samples, the response yielding the highest $R_{BEACON}$ score is selected as $y_w$, and the one with the lowest $R_{BEACON}$ score is selected as $y_l$. The same discard rule applies if all $K$ responses result in identical $R_{BEACON}$ scores.

\subsubsection{Iterative Direct Preference Optimization Training}
\label{app:dpo_training}
The policy model, Qwen2.5-7B-instruct, is fine-tuned using Direct Preference Optimization (DPO) \citep{Rafailov2023DirectPO} on the preference pairs $(y_w, y_l)$ generated by either the BoN or BEACON strategy. Key hyperparameters for DPO training include a learning rate of $5 \times 10^{-7}$, a global batch size of 128, and a maximum sequence length of 4096. In each DPO iteration, the policy model is trained for 2 epochs on the newly generated preference dataset. The entire cycle of preference pair generation (using either BoN or BEACON) followed by DPO fine-tuning of the policy model is repeated for a total of 7 iterations.

\subsubsection{Reward Adaptation for BEACON Algorithm}
\label{sssec:reward_adaptation_beacon} 
The BEACON algorithm, particularly its Bayesian parameter updates, expects a continuous reward signal to estimate the underlying reward distribution effectively. While our foundational reward assessment might involve rule-based, binary outcomes (e.g., correct/incorrect), we adapt the score $R_{learned}$ from our fixed reward model (described in Appendix~\ref{app:fixed_reward_model}) to better suit BEACON's requirements. This adaptation is based on the correctness of the final answer found within a \verb|\boxed{}| environment in the generated response.

Let $R_{learned}$ be the continuous score produced by our fixed reward model for a given response. The final reward, $R_{BEACON}$, used by the BEACON algorithm is determined as follows:
\begin{itemize}
    \item If the response contains the correct final answer in a \verb|\boxed{}| environment:
        \begin{itemize}
            \item If $R_{learned} > 0$, then $R_{BEACON} = R_{learned} \times 2$.
            \item If $R_{learned} \le 0$, then $R_{BEACON} = R_{learned} / 2$ (making a non-positive score less detrimental if the final answer is surprisingly correct).
        \end{itemize}
    \item If the response contains an incorrect final answer in a \verb|\boxed{}| environment:
        \begin{itemize}
            \item If $R_{learned} > 0$, then $R_{BEACON} = R_{learned} / 2$ (penalizing a score that might have seemed good otherwise).
            \item If $R_{learned} \le 0$, then $R_{BEACON} = R_{learned} \times 2$ (further penalizing an already non-positive score).
        \end{itemize}
    \item If the response fails to provide any final answer in a \verb|\boxed{}| environment, the reward remains unchanged: $R_{BEACON} = R_{learned}$.
\end{itemize}

\section{Additional Theoretical Analysis and Proofs} 
This section provides supplementary theoretical background and proofs for the BEACON framework.

\subsection{Classical Sequential Search Problem}\label{app:sequential_search}
\begin{figure*}[h!]
    \centering
    \includegraphics[width=0.6\textwidth]{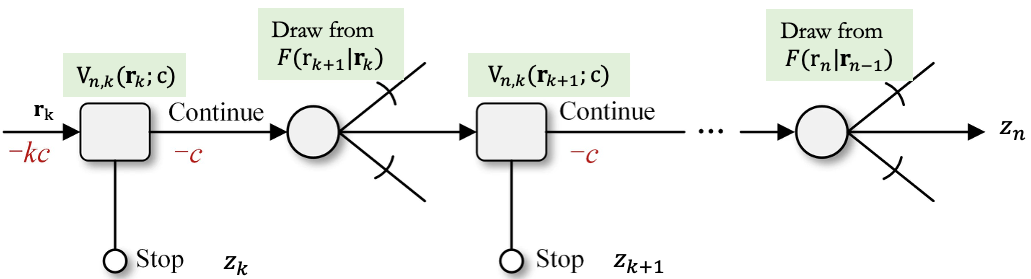}
    \caption{Sequential search problem illustration: At each step, the decision-maker observes a reward $r_k$ and decides whether to stop with the current best reward $z_k$ or continue sampling, incurring cost $c$.}
    \label{fig:sequential_search}
\end{figure*}

In the classical sequential search problem \citep{weitzman1979optimal}, a decision maker (DM) faces an infinite sequence of independent offers $\{x_1,x_2,\dots\}$ drawn from a known distribution function $F$ with density $f$. Sampling incurs a constant cost $c>0$, and the DM may stop at any time, accepting the highest offer observed so far (see Figure~\ref{fig:sequential_search} for illustration). Since the distribution $F$ is fully known, the predictive distribution remains unchanged throughout the process.

The dynamic programming recursion for the value function after $k$ samples is:
\[
V(z) = \max\Bigg\{ z,\ \int_{-\infty}^\infty V\left(\max\{z,y\}\right) dF(y) - c \Bigg\}
\]
where $z$ is the current best observed offer. The associated gain from one more search is $H(z) = \int_z^\infty (y-z) dF(y),$
representing the expected improvement conditional on continuing.

A fundamental property is the **reservation price property**: there exists a threshold $r^\ast$ such that the optimal policy is to stop if and only if $z \geq r^\ast$. This reservation price solves:
\[
c = H(r^\ast) = \int_{r^\ast}^\infty (y-r^\ast) dF(y).
\]

The DM thus compares the sampling cost with the expected marginal benefit; once the best offer reaches $r^\ast$, the expected gain no longer justifies continued search.

This threshold-based policy has important implications: (1) $r^\ast$ depends only on $F$ and $c$, not on the number of samples or their sequence; (2) as $c$ increases, $r^\ast$ decreases, leading to earlier stopping; and (3) distributions with heavier right tails yield higher reservation prices, reflecting greater potential benefits from continued search.

\subsection{Priors and Minimal Sample Size for Bayesian Updating}
\label{app:minimal_sample_size}
We employ a conjugate Normal-Inverse-Gamma prior for the unknown mean $\mu$ and variance $\sigma^2$ of the reward distribution, following \citep{baucells2024search}. In this framework, the precision $1/\sigma^2$ follows a Gamma distribution with parameters $(\alpha_0, \beta_0)$, and conditional on precision, $\mu$ follows a Normal distribution with mean $\mu_0$ and variance $\sigma^2/\nu_0$. This structure enables analytical posterior updates with hyperparameters $(\alpha_k, \nu_k, \beta_k, \mu_k)$.
Starting with prior NIG$(\mu_0, \nu_0, \alpha_0, \beta_0)$ and after observing $k$ samples, the posterior parameters become:
\begin{align}\label{eq:hyperamareters_update}
\begin{aligned}
&\alpha_k = \alpha_0 + \frac{k}{2}, \nu_k = \nu_0 + k,  \mu_k = \frac{\nu_0\mu_0 + k\bar{r}_k}{\nu_0 + k}, \\
&\beta_k = \beta_0 + \frac{\sum_{i=1}^k(r_i-\bar{r}_k)^2}{2} + \frac{k\nu_0(\bar{r}_k-\mu_0)^2}{2(\nu_0+k)},
\end{aligned}
\end{align}
where $\bar{r}_k$ is the sample mean. The posterior predictive distribution follows a Student-t distribution with $2\alpha_k$ degrees of freedom, mean $\mu_k$, and scale parameter $\sigma_k = \sqrt{(\nu_k+1)\beta_k/(\nu_k\alpha_k)}$.

\begin{table}[h]
\footnotesize
\centering
\caption{Prior configurations and their associated minimum sample size $k_0$ required for a well-defined posterior predictive distribution.}
\begin{tabular}{| l c |} 
\hline
\textbf{Prior Configuration} & \textbf{$k_0$} \\
\hline
$2\alpha_0> 1$, $\nu_0 ,\beta_0 > 0$ & 0 \\
$2\alpha_0 \in (0, 1]$, $\nu_0 > 0$, $\beta_0 \geq 0$ & 1 \\
$2\alpha_0> 1$, $\nu_0 > 0$, $\beta_0 = 0$ & 1 \\
$\alpha_0,\beta_0> 0$, $\nu_0 = 0$ & 1 \\
$\alpha_0> 0$, $\nu_0=\beta_0=0$ & 2 \\
$2\alpha_0 \in (-1, 0]$, $\nu_0 ,\beta_0 \geq 0$ & 2 \\
$2\alpha_0=-1$, $\nu_0 ,\beta_0 \geq 0$ & 3 \\
\hline
\end{tabular}
\label{table_k0}
\end{table}

The minimal sample size $k_0$ depends on the chosen prior hyperparameters. For BEACON, we adopt Jeffreys' non-informative prior $(\alpha_0, \nu_0, \beta_0)=(-1/2, 0, 0)$, which requires $k_0=3$ initial observations before a valid posterior predictive distribution emerges. More informative priors require fewer initial samples, as summarized in Table \ref{table_k0}.

\subsection{Sufficient Statistics for the Search Problem using Bayesian Updating}
\label{app:sufficent_statistics}
The Bayesian sequential search model can be significantly simplified through sufficient statistics that encapsulate all relevant information for optimal stopping decisions \citep{baucells2024search}.

\begin{lemma}
 For any sampling stage $k$, where $k_0 \leq k \leq n$, the triple $(z_k,\mu_k,\sigma_k)$ together with $k$ constitute sufficient statistics for the value-to-go function $V_{n,k}(\mathbf{r}_k;c)$.
\end{lemma}

This lemma establishes that rather than tracking the complete observation history $\mathbf{r}_k = \{r_1,\ldots,r_k\}$, we only need to maintain three key statistics:
\begin{itemize}
    \item $z_k=\max\{r_1,\dots,r_k\}$: The highest observed reward so far
    \item $\mu_k$: The posterior mean of the reward distribution
    \item $\sigma_k$: The scale parameter derived from the posterior distribution
\end{itemize}

These statistics update efficiently according to the following formula when observing a new reward $r_{k+1}$:
\begin{align}\label{parameter_update}
z_{k+1} &= \max\{z_k, r_{k+1}\}, \\
\mu_{k+1} &= \mu_k + \frac{r_{k+1}-\mu_k}{\nu_0+k+1}, \\
\sigma_{k+1} &= 
\sqrt{\frac{1-\tfrac{1}{(\nu_0+k+1)^2}}{\,2\alpha_0+k+1\,}} \nonumber \\
&\quad \times \sqrt{(2\alpha_0+k)\sigma_k^2 + (r_{k+1}-\mu_k)^2}.
\end{align}

The posterior predictive distribution follows a Student-t distribution with $2\alpha_k$ degrees of freedom. This statistical sufficiency substantially simplifies both theoretical analysis and practical implementation of BEACON by reducing the problem's dimensionality from tracking $k$ individual rewards to monitoring just three informative statistics, making the algorithm computationally efficient and mathematically tractable.

\subsection{Details of Universal Index Policy}\label{app:UIP}
Here we elaborate on the Universal Index Policy.

\begin{definition}\label{def_1} For $k_0\leq k< n$, the \textit{index function} $h_{n,k}:\mathbb{R}\to (0,\infty)$ maps each standardized best reward $\hat{z}\in \mathbb{R}$ to the unique value $c>0$ that solves $H_{n,k}(\hat{z},0,1;c)=c$. 
\end{definition}

This definition captures how the h-index represents the exact cost threshold where the expected value of continuing equals that of stopping. The strength of this approach is its reusability—once computed, these indices apply across different queries with similar statistical properties.

\begin{theorem} \label{UIP}\citep{baucells2024search} After $k\geq k_0$ observations, with standardized best reward $\hat{z}_k=(z_k-\mu_k)/\sigma_k$, it is optimal to stop and accept $z_k$ if $c\geq\sigma_kh_{n,k}(\hat z_k)$, and continue searching otherwise. Equivalently, stop if and only if~$\hat{z}_k \geq h^{-1}_{n,k}(c/\sigma_k)$.
\end{theorem}

\begin{theorem}\citep{baucells2024search} \label{h_properties}
The h-index function $h_{n,k}(\hat{z})$ for $k_0 \leq k< n$ is strictly decreasing, convex, and satisfies $\lim_{\hat{z}\rightarrow\infty} h_{n,k}(\hat{z})=~0$.
\end{theorem}

This theorem establishes BEACON's core stopping criterion—after normalizing the current best reward, we stop sampling when the cost-adjusted index threshold is reached. This occurs when the current best reward is sufficiently high relative to our uncertainty about the reward distribution, considering the sampling cost.

\subsection{Computation of the h-index}\label{app:compute_h_index}
Computing the h-index function $h_{n,k}(\hat{z})$ requires solving recursive equations based on the Bellman equation and expected marginal gain function $H_{n,k}$ from \S\ref{sec:model}. As derived in \citep{baucells2024search}, we need to solve:
\begin{align}\label{eq:compute_h_index}
H_{n,k}(&\hat{z},0,1;c) = H_{k+1,k}(\hat{z},0,1;c) \nonumber \\
&+ \int \sigma_u \cdot g(u) \, dF_{2\alpha}(u),
\end{align}
where
$$
g(u) = \max\bigg\{ 0, H_{n,k+1}\bigg(\frac{z_u-\mu_u}{\sigma_u},0,1;\frac{c}{\sigma_u}\bigg)\bigg\}
$$

with boundary condition $H_{n,n}=0$, where $z_u =\max\{\hat{z},u\}$, $\mu_u = u / (\nu_0+k+1)$, $\sigma_u = \Lambda_{k+1} \sqrt{2\alpha_0+k +u^2}$, $\Lambda_{k+1} = \sqrt{\frac{2\alpha_0+k}{2\alpha_0+k+1}}$, and $F_{2\alpha}$ is the CDF of the Student-t distribution with $2\alpha_0$ degrees of freedom. 

In our implementation, we pre-compute the h-index function $h_{n,k}(\hat{z})$ for each time horizon $n$ using Jeffreys' non-informative prior ($\alpha_0=-0.5,\nu_0=0$). We create a lookup table using a geometric grid of $\hat{z}_k \in [-30,30]$ with resolution $G=100$ for all timesteps $k \in [3,n]$, leveraging an optimized implementation\footnote{\url{https://github.com/MSORlearners/h-index-computation.git}} with Numba for parallel processing.

During inference, BEACON evaluates the stopping condition through constant-time linear interpolation on this pre-computed table, enabling efficient decision-making during sequential sampling without the computational overhead of dynamic programming calculations at runtime.

\subsection{Binary-Reward Variant: Bernoulli--Beta Learning}\label{app:discrete_case}

\newcommand{\q}[2]{#1/(#1{+}#2)}

\paragraph{Model.}
Consider the sequential generation process in Section~\ref{sec:model} with binary rewards $r_i\in\{0,1\}$ and unknown success probability $\theta\in(0,1)$. Conditional on $\theta$, the draws are i.i.d.\ $\mathrm{Bernoulli}(\theta)$. Sampling incurs a per-draw cost $c>0$ and the horizon is at most $n$ draws. Let $\mathbf r_k=(r_1,\dots,r_k)$ and $z_k=\max\{r_1,\dots,r_k\}\in\{0,1\}$ be the best observed reward by stage $k$. Adopt the conjugate prior $\theta\sim\mathrm{Beta}(a_0,b_0)$ with $a_0,b_0>0$. After $k$ draws with $s_k=\sum_{i=1}^k r_i$ successes and $f_k=k-s_k$ failures, the posterior is
\[
\textstyle
\theta\mid \mathbf r_k \sim \mathrm{Beta}(a_k,b_k),\quad
a_k=a_0+s_k,\ \ b_k=b_0+f_k,
\]
and the posterior-predictive probability of success on the next draw is
\[
\textstyle
q_k = \Pr(r_{k+1}=1\mid \mathbf r_k)=\q{a_k}{b_k}.
\]
The sufficient statistics are $(z_k,a_k,b_k)$. Upon observing $r_{k+1}$,
\begin{align}\label{eq:bernoulli-update}
&z_{k+1}=\max\{z_k,r_{k+1}\},\\
&a_{k+1}=a_k+r_{k+1}, b_{k+1}=b_k+1-r_{k+1}.
\end{align}

\paragraph{Dynamic program.}
If $z_k=1$ the process is absorbing and the decision maker stops with value $1$. When $z_k=0$, let $V_{n,k}(0,a_k,b_k;c)$ denote the optimal value (with at most $n$ draws total, at stage $k$). Then
\begin{align}\label{eq:bellman-bernoulli}
V_{n,k}(0,a_k,b_k;c) &= \max\{0,v_k(c)\}, \\
V_{n,n}(0,a_n,b_n;c) &= 0, \quad
V_{n,k}(1,a_k,b_k;c) = 1,
\end{align}
where $v_k(c)= q_k + (1-q_k)\,V_{n,k+1}\bigl(0,a_k,b_k+1;c\bigr) - c$, 
and $q_k=\q{a_k}{b_k}$.

It is convenient to rewrite the recursion in remaining-draws form. Let $R_t(a,b;c)$ be the optimal value with $t\in\{0,1,\dots\}$ draws remaining and no success yet. With $R_0(\cdot)\equiv 0$,
\begin{align}\label{eq:R-recursion}
\begin{aligned}
    \textstyle
R_t&(a,b;c)=\max\bigl\{0,q(a,b)+\\
&(1-q(a,b))\,R_{t-1}(a,b+1;c)-c\bigr\},
\end{aligned}
\end{align}
where $q(a,b)=\q{a}{b}$. The connection to \eqref{eq:bellman-bernoulli} is $V_{n,k}(0,a_k,b_k;c)=R_{n-k}(a_k,b_k;c)$.

\paragraph{Reservation cost and optimal policy.}
For each stage, there is a unique cost threshold at which the decision maker is indifferent between stopping and taking one more draw.

\begin{lemma}[Basic properties of $R_t$]\label{lem:properties}
For fixed $(t,a,b)$ with $t\ge1$ and $a,b>0$, the map $c\mapsto R_t(a,b;c)$ is continuous, nonincreasing, and $1$-Lipschitz on $[0,\infty)$. Moreover $R_t(a,b;0)>0$ and $R_t(a,b;1)=0$; hence $\{c>0:R_t(a,b;c)>0\}\subset(0,1)$.
\end{lemma}

\textbf{Proof: }
Induct on $t$. The base $t=0$ is trivial since $R_0\equiv0$. Suppose the claim holds for $t-1$. The inner term
\[
\textstyle
\psi_t(c)\coloneqq q(a,b)+(1-q(a,b))\,R_{t-1}(a,b+1;c)-c
\]
is continuous and $1$-Lipschitz as a sum of a constant, a nonnegative multiple of a $1$-Lipschitz function, and $-c$. The outer map $x\mapsto \max\{0,x\}$ is continuous and $1$-Lipschitz, hence so is $R_t=\max\{0,\psi_t\}$. Monotonicity in $c$ follows from the $-c$ term. For $c=0$, $R_t(a,b;0)\ge R_1(a,b;0)=q(a,b)>0$. For $c=1$, since $R_{t-1}(a,b+1;1)\le R_{t-1}(a,b+1;0)$ and $q(a,b)\le 1$, one has $\psi_t(1)\le q(a,b)-1\le0$, hence $R_t(a,b;1)=0$.
\qed

\begin{lemma}[Strictly decreasing “continue” margin]\label{lem:strict}
Define $\phi_t(c)\coloneqq q(a,b)+(1-q(a,b))\,R_{t-1}(a,b+1;c)-c$. Then $\phi_t$ is continuous and strictly decreasing on $[0,\infty)$, with $\phi_t(0)>0$ and $\phi_t(1)\le0$.
\end{lemma}

\textbf{Proof: }
Continuity follows from Lemma~\ref{lem:properties}. If $0\le c_0<c_1$, then using that $R_{t-1}$ is nonincreasing and $1$-Lipschitz,
\begin{align*}
\phi_t(c_1)-\phi_t(c_0)&=(1-q)\bigl(R_{t-1}(c_1)-R_{t-1}(c_0)\bigr)\\
&-(c_1-c_0)\le 0-(c_1-c_0)<0,
\end{align*}
so $\phi_t$ is strictly decreasing. The sign claims are from Lemma~\ref{lem:properties}.
\qed

\begin{theorem}[Reservation--cost policy]\label{thm:bernoulli_policy}
Fix $n$ and a stage $k<n$ with posterior parameters $(a_k,b_k)$. There exists a unique reservation cost $h^{\mathrm B}_{n,k}(a_k,b_k)\in[0,1]$ such that
\[
\textstyle
V_{n,k}(0,a_k,b_k;c)>0 \iff c<h^{\mathrm B}_{n,k}(a_k,b_k).
\]
Equivalently, $h^{\mathrm B}_{n,k}(a_k,b_k)$ is the unique solution $c$ to
\begin{equation}\label{eq:reservation-fixed-point}
\textstyle
c = \tfrac{a_k}{a_k+b_k} + \tfrac{b_k}{a_k+b_k} V_{n,k+1}\bigl(0,a_k,b_k+1;\,c\bigr).
\end{equation}
The Bayes-optimal policy is: if $z_k=1$, stop and obtain $1$; if $z_k=0$, continue iff $c<h^{\mathrm B}_{n,k}(a_k,b_k)$.
\end{theorem}

\textbf{Proof: }
Work with $R_t$ at $t=n-k$. By Lemma~\ref{lem:strict}, $\phi_t$ is strictly decreasing and continuous with $\phi_t(0)>0$ and $\phi_t(1)\le0$, hence it has a unique root $h_t(a,b)\in[0,1]$. From \eqref{eq:R-recursion}, $R_t=\max\{0,\phi_t\}$, so $R_t>0$ iff $c<h_t(a,b)$. Continuity gives $\phi_t(h_t)=0$, which is \eqref{eq:reservation-fixed-point} after mapping $(t,a,b)$ back to $(n,k,a_k,b_k)$. The stated action rule is exactly the comparison between the two branches in \eqref{eq:bellman-bernoulli}, while $z_k=1$ is absorbing with value $1$.
\qed
\textbf{Adaptation to BEACON:} For binary rewards, BEACON replaces the continuous Normal framework with Bernoulli-Beta conjugate learning. The stopping criterion from Theorem \ref{thm:optimal_policy} becomes the binary reservation cost policy in Theorem \ref{thm:bernoulli_policy}, while the sufficient statistics \eqref{parameter_update} are replaced with the simpler Beta updates \eqref{eq:bernoulli-update}. This maintains BEACON's sequential structure while using binary-specific optimal stopping decisions.

\subsection{Proof of Theorem \ref{thm:optimal_policy}}\label{app:proof_optimal_policy}
\textbf{Proof.}
Under the Normal reward model with Normal–Inverse–Gamma prior, $(z_k,\mu_k,\sigma_k)$ (together with $k$) are sufficient statistics for the state (Lemma in Appendix \ref{app:sufficent_statistics}). Standardizing gives $\hat z_k=(z_k-\mu_k)/\sigma_k$ and reduces the problem to the canonical (location–scale normalized) sequential search instance.

By Theorem \ref{UIP} (Universal Index Policy), for the canonical problem there exists a strictly decreasing index function $h_{n,k}(\cdot)$ such that the (myopic) continuation rule
\[
\text{Continue at step }k \quad \Longleftrightarrow \quad h_{n,k}(\hat z_k)>\frac{c}{\sigma_k}
\]
maximizes the Bellman value function. Translating back to original (unscaled) variables yields exactly condition \eqref{eq:UIP}. Therefore the stopping time
\[
K=\min\{k\ge k_0: h_{n,k}(\hat z_k)\le c/\sigma_k\}\wedge n
\]
achieves the optimal value $\mathbb{E}[z_K-Kc]$.

Optimality of $K$ follows because: (i) any earlier stop with $h_{n,k}(\hat z_k)>c/\sigma_k$ forgoes strictly positive expected marginal gain; (ii) any continuation with $h_{n,k}(\hat z_k)\le c/\sigma_k$ incurs cost exceeding expected benefit; (iii) strict monotonicity of $h_{n,k}$ implies no alternative rule can dominate. Hence $K$ is the unique optimal stopping time under the stated assumptions. \qed

\subsection{Sensitivity Analysis}
\label{app: prop1}
\begin{proposition}\label{proposition_1}
The optimal stopping time $K$ under the policy is influenced by several factors: it decreases with higher sampling cost $c$ and larger current best reward $z_k$, while increasing with higher posterior mean $\mu_k$ and greater posterior scale parameter $\sigma_k$. Additionally, the algorithm becomes more patient (tend to continue) when more remaining samples ($n-k$) are available.
\end{proposition}

\textbf{Proof of Proposition \ref{proposition_1}}\label{app: prop1}
\textbf{Proof: } From Theorem \ref{h_properties}, we know that the h-index function $h_{n,k}(\hat{z})$ for $k_0 \leq k< n$ is strictly decreasing, convex, and satisfies $\lim_{\hat{z}\rightarrow\infty} h_{n,k}(\hat{z})=~0$.

Under the UIP stopping criterion in Equation \eqref{eq:UIP}, stopping occurs when $h_{n,k}(\hat{z}_k) \leq \frac{c}{\sigma_k}$. Since $h_{n,k}(\hat{z})$ is strictly decreasing, higher $c$ raises the threshold $h^{-1}_{n,k}(c/\sigma_k)$, leading to earlier stopping (smaller $K$). 

For a fixed standardized best reward $\hat{z}$, the h-index function $h_{n,k}(\hat{z})$ is increasing in $n-k$ (the number of remaining samples) as shown in \citet{baucells2024search}. This means that when more samples remain available (larger $n-k$), the threshold for stopping becomes higher, making the algorithm more "patient" and likely to continue sampling.

With $\hat{z}_k = \frac{z_k - \mu_k}{\sigma_k}$, higher $z_k$ increases $\hat{z}_k$, which decreases $h_{n,k}(\hat{z}_k)$ due to the function's monotonicity, resulting in earlier stopping. Conversely, higher $\mu_k$ decreases $\hat{z}_k$, thereby increasing $h_{n,k}(\hat{z}_k)$ and extending sampling (larger $K$). The scale parameter $\sigma_k$ affects stopping through dual mechanisms: it decreases $\hat{z}_k$ by appearing in the denominator and simultaneously decreases $\frac{c}{\sigma_k}$. Both effects increase $h_{n,k}(\hat{z}_k)$ relative to $\frac{c}{\sigma_k}$, encouraging continued sampling (larger $K$).

\section{Additional Experimental Results}

\subsection{Statistical Significance Analysis of BEACON}
\label{app:error_bar}

\newcommand{\bluepm}[1]{\textcolor{blue}{$\pm$#1}}

\begin{table*}[ht]
\centering
\caption{Comparison of BEACON with baseline methods using LLaMA-3.2-3B. Results shown as mean\bluepm{SEM} across 5 random seeds.}
\label{tab:llama_significance}
\small  
\renewcommand{\arraystretch}{1.3}
\setlength{\tabcolsep}{4pt}
\begin{tabular}{@{}l c c c c c c c c@{}}
\toprule
 & \multicolumn{4}{c}{\textbf{Reasoning Tasks (Avg. MATH/AIME/AMC)}} & \multicolumn{4}{c}{\textbf{Alignment Task (AlpacaEval 2.0)}} \\
\cmidrule(lr){2-5} \cmidrule(lr){6-9}
\textbf{Method} & \textbf{Acc.~↑} & \textbf{Samples~↓} & \textbf{Reward~↑} & \textbf{Value~↑} & \textbf{Win~↑} & \textbf{Samples~↓} & \textbf{Reward~↑} & \textbf{Value~↑} \\
 & \textbf{(\%)} & \textbf{($\overline{K}$)} & \textbf{(Scaled)} & \textbf{(Scaled)} & \textbf{(\%)} & \textbf{($\overline{K}$)} & \textbf{(Scaled)} & \textbf{(Scaled)} \\
\midrule
Direct CoT & 20.0\bluepm{2.8} & 1.0 & $-$1.60\bluepm{0.10} & $-$0.40\bluepm{0.05} & 16.0\bluepm{3.5} & 1.0 & 0.20\bluepm{0.05} & $-$0.80\bluepm{0.08} \\
BoN (N-RM) & 33.4\bluepm{1.3} & 32.0 & 3.49\bluepm{0.22} & 0.29\bluepm{0.14} & 25.0\bluepm{2.5} & 32.0 & 4.00\bluepm{0.25} & 0.80\bluepm{0.09} \\
\rowcolor{lightgray!30}\textbf{BEACON (N-RM)} & 32.8\bluepm{1.6} & 15.8\bluepm{1.2} & 3.25\bluepm{0.18} & \textbf{1.12}\bluepm{0.21} & 23.5\bluepm{1.8} & 14.5\bluepm{2.3} & 3.65\bluepm{0.22} & \textbf{1.20}\bluepm{0.20} \\
\bottomrule
\end{tabular}
\end{table*}

To assess the reliability of our results, we present a focused analysis using LLaMA-3.2-3B as our base model. Each experiment was conducted with 5 different random seeds, and we report the error bars as the standard error of the mean (SEM). As shown in Table~\ref{tab:llama_significance}, BEACON achieves comparable performance to the BoN baseline in terms of accuracy (32.8\bluepm{1.6}\% vs. 33.4\bluepm{1.3}\% for reasoning tasks) and win rate (23.5\bluepm{1.8}\% vs. 25.0\bluepm{2.5}\% for alignment tasks), with overlapping error margins indicating no substantial performance degradation. However, BEACON requires significantly fewer samples (15.8\bluepm{1.2} vs. 32.0 for reasoning tasks), resulting in substantially higher value scores that account for both performance and computational cost (1.12\bluepm{0.21} vs. 0.29\bluepm{0.14}).

\subsection{Impacts of Cost on the Value Optimizations}
\label{app:cost_v}

The sampling cost, denoted by $c$, plays a critical role in the optimization of our value function, which serves as the core objective for determining the optimal stopping criterion. As illustrated in Figure~\ref{fig:cost_analysis_value_functions}, variations in the cost parameter directly influence the shape and peak of the value function and the resulting optimal sample size. The left subplot of Figure~\ref{fig:cost_analysis_value_functions} shows that for any given sampling cost, the value function initially increases with the number of samples as the expected reward grows when we have more sample option to select, but eventually decreases as the cumulative cost outweighs the marginal gain from additional samples. Notably, increasing the sampling cost leads to a lower maximum achievable value and shifts the point of optimal stopping (where the value function peaks) towards a smaller number of samples. The right subplot of Figure~\ref{fig:cost_analysis_value_functions} further emphasizes this relationship by directly plotting the optimal sample size against the sampling cost. This plot clearly demonstrates a strong inverse correlation: as the cost of obtaining each sample increases, the BEACON framework optimally decides to stop sampling earlier, resulting in a significantly reduced optimal sample size.

\subsection{Normality Analysis of Reward Distributions}
\label{app:norm_analysis}

While BEACON leverage learning for reward estimation, we acknowledge that real-world reward distributions from LLMs may not always strictly adhere to normality. This appendix section visually explores the characteristics of reward distributions observed in our experiments using the Nemotron reward model, providing context for our robust updating mechanism.

\begin{figure}[htb!] 
    \centering
    \includegraphics[width=0.8\linewidth]{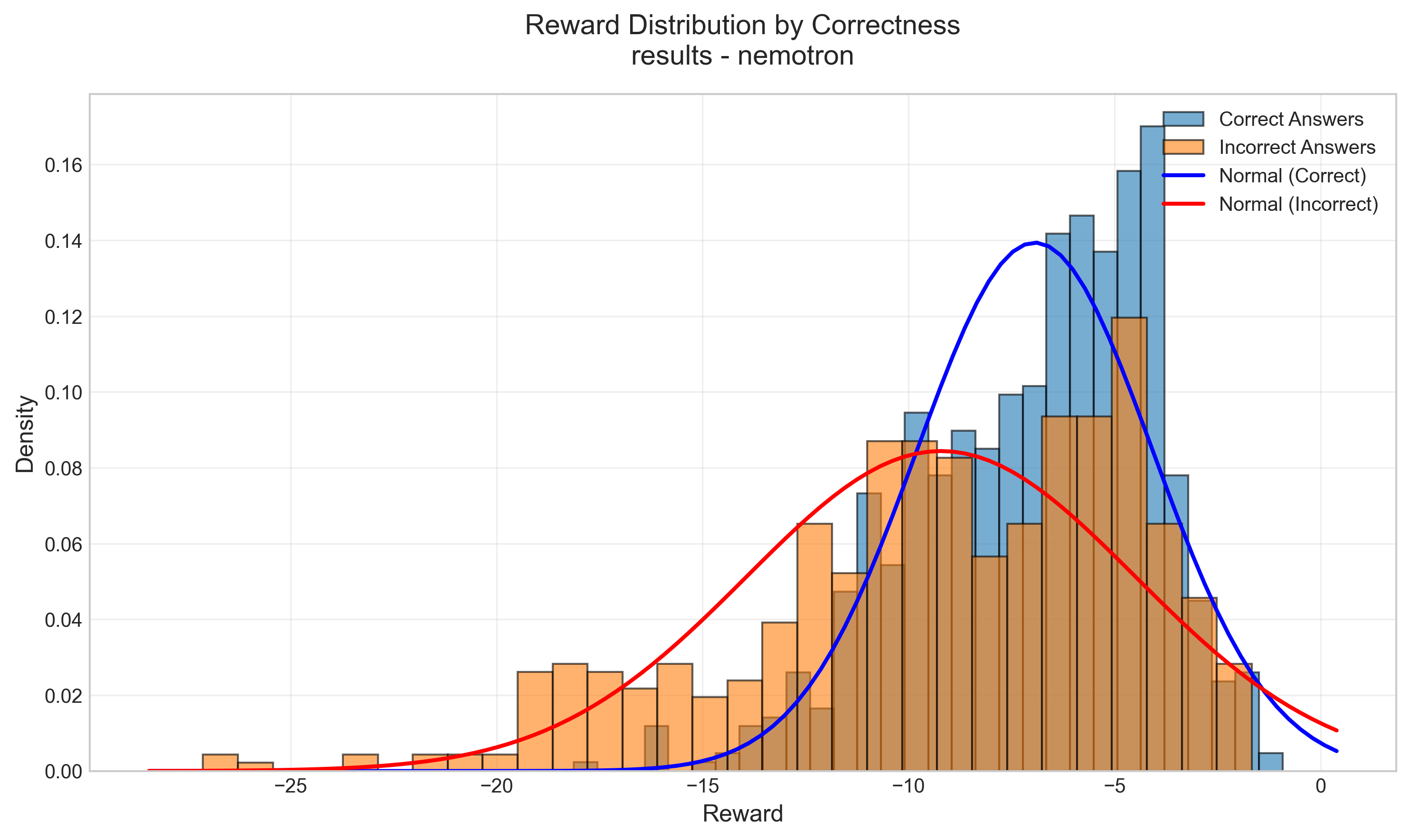} 
    \captionof{figure}{Aggregated reward distributions from the Nemotron RM, separated for responses classified as correct (blue histogram, blue normal fit) and incorrect (orange histogram, red normal fit). Both distributions show approximate normality.}
    \label{fig:normality_corr}
\end{figure}

Figure~\ref{fig:normality_corr} presents an aggregated view of reward distributions, conditioned on whether the generated responses were ultimately deemed correct or incorrect. We observe that for both categories, the empirical distributions of rewards are reasonably approximated by a normal distribution, albeit with different means and variances. Specifically, correct answers tend to receive higher mean rewards, but both distributions exhibit a unimodal, bell-like shape characteristic of normality. This overall trend provides a foundational justification for employing a Gaussian-based learning model.

\begin{figure}[htb!] 
    \centering
    \includegraphics[width=\linewidth]{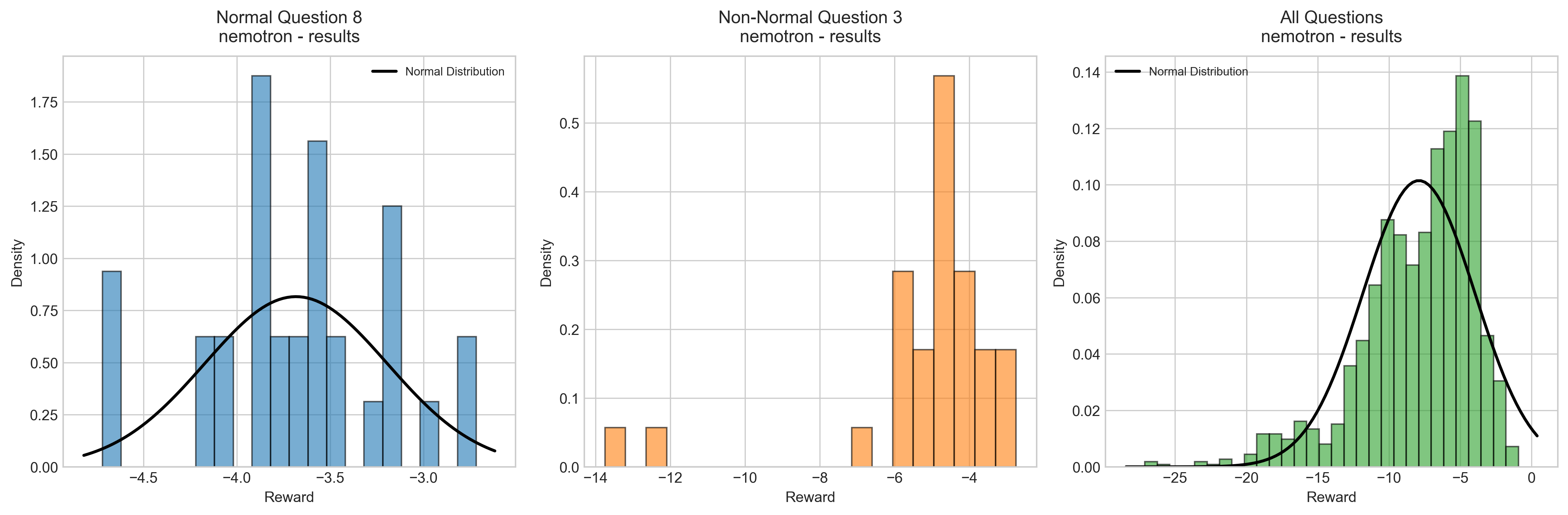} 
    \captionof{figure}{Examples of query-specific reward distributions using the Nemotron RM. Left: A typical case exhibiting approximate normality ("Normal Question 8"). Middle: An occasional case with predominantly high rewards and some low-value outliers ("Non-Normal Question 3"), motivating robust updates. Right: Aggregate distribution of all rewards.}
    \label{fig:normality_case}
\end{figure}

However, analyzing distributions at an aggregate level can mask variations in individual query-specific reward patterns. Therefore, Figure~\ref{fig:normality_case} dives into specific cases to illustrate the types of reward distributions encountered for individual prompts. The leftmost panel ("Normal Question 8") depicts a common scenario where the rewards for multiple samples from a single prompt follow an approximately normal distribution, though the specific mean and variance naturally differ from prompt to prompt. In contrast, the middle panel ("Non-Normal Question 3") illustrates an occasional but important pattern: the distribution consists primarily of high-reward samples with a few significantly lower, noisy rewards in the left tail. This type of skewed distribution, or one with outliers, can badly influence standard posterior parameter updates. But it is precisely these instances that motivate BEACON's robust updating formula, which is designed to mitigate the impact of such extreme low-value outliers, thereby maintaining a more stable and reliable estimation of the reward potential focused on the right tail. The rightmost panel ("All Questions") shows the overall distribution of all rewards for context.

These observations support our approach: while normality is a useful working assumption for the bulk of reward behaviors, the adaptive robust update mechanism provides resilience against deviations, particularly those caused by uninformative low scores, ensuring BEACON remains effective across diverse and sometimes non-ideal reward landscapes.

\subsection{Robust Updating Formula and Details}
\label{app:robust_update}

\paragraph{Robust Update Rule.}  
To mitigate negative skewness and extreme left-tail outliers, we modify the standard posterior update by filtering rewards below the 1\% posterior-predictive quantile. Specifically,  
\begin{align}
z_{k+1} &= \max\{z_k, r_{k+1}\}, \\
\tilde{r}_{k+1} &= 
\begin{cases}
\mu_k, & \text{if } r_{k+1} < q_{0.01}, \\
r_{k+1}, & \text{otherwise},
\end{cases} \\
\mu_{k+1} &= \mu_k + \frac{\tilde{r}_{k+1}-\mu_k}{\nu_0+k+1}, \\
\sigma_{k+1} &= 
\sqrt{\frac{1-\tfrac{1}{(\nu_0+k+1)^2}}{\,2\alpha_0+k+1\,}} \nonumber \\
&\quad \times \sqrt{(2\alpha_0+k)\sigma_k^2 + (\tilde{r}_{k+1}-\mu_k)^2}.
\end{align}

where $q_{0.01} = F^{-1}_{2\alpha_k}(0.01 \mid \mu_k, \sigma_k)$ is the $1\%$ quantile of the posterior predictive distribution.  

\paragraph{Interpretation.}  
This one-sided winsorization caps the influence of extreme left-tail samples at the posterior mean, reducing variance inflation while leaving the maximum statistic $z_{k+1}$ intact unless a new best reward is observed. The adjustment preserves the right-tail fidelity of the reward distribution, which is essential for correctly identifying high-quality responses.

\paragraph{Practical Notes.}  
(i) The choice of threshold $p$ is robust across $[0.5\%,2\%]$, with $p=1\%$ as default.  
(ii) The update is $O(1)$ per step; quantiles can be pre-tabulated for efficiency.  
(iii) For numerical stability, enforce $\sigma_k \geq 10^{-6}$.  
(iv) Optionally, robust updates can be activated only when recent empirical skewness is strongly negative (e.g., $\gamma_1 < -0.5$).

\paragraph{Results.}

\begin{figure}[t]
  \centering
  \includegraphics[width=0.88\linewidth]{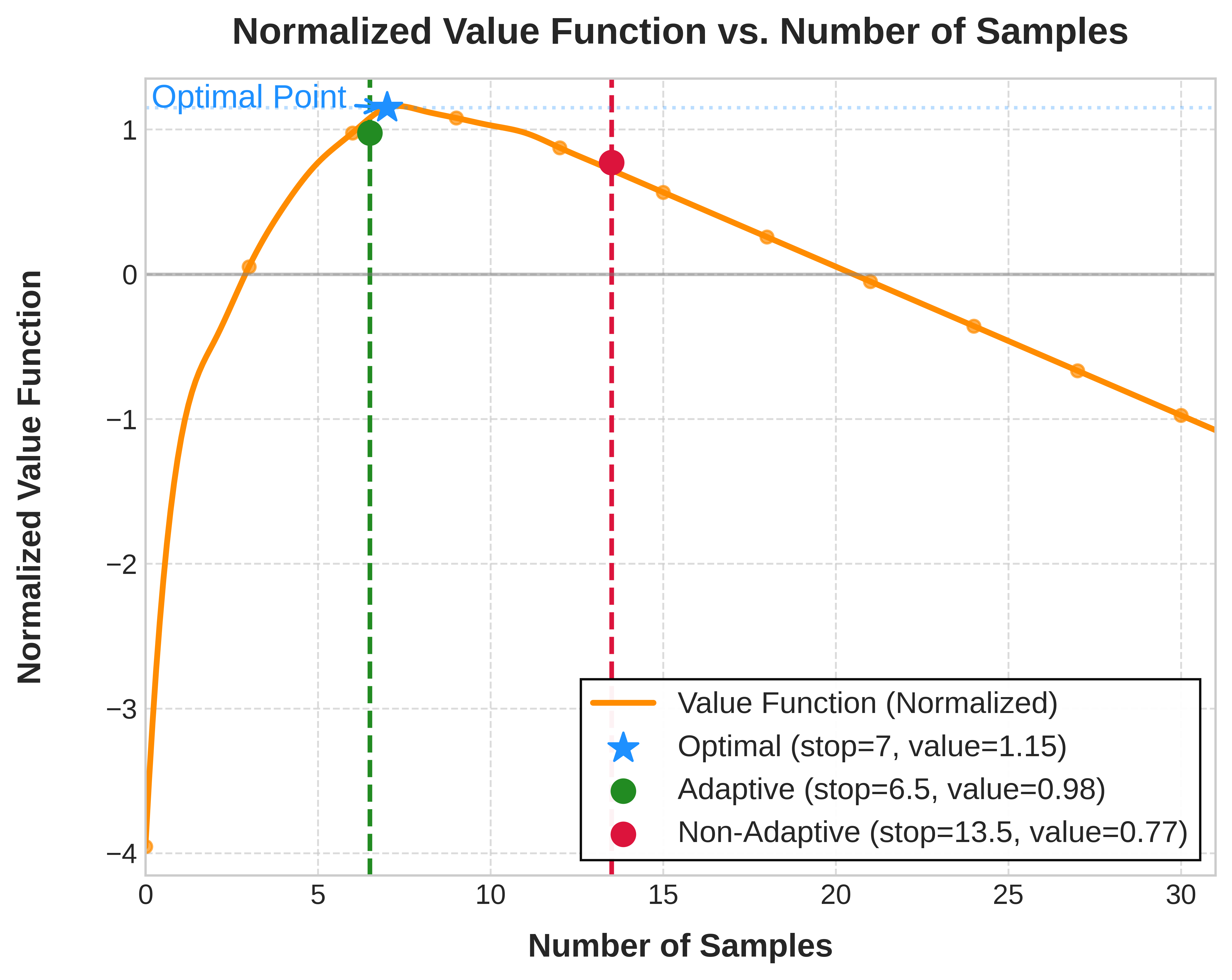}
  \caption{Value Estimation for the robust parameter update method (adaptive) vs.~non-adaptive method. Our design helps BEACON avoid violating assumptions and stop closer to the optimum on average.}
  \label{fig:side_by_side_adaptive_plot}
  \vspace{-8pt}
\end{figure}

\begin{table*}[htbp]
\centering
\caption{Examples of BEACON Behavior and Sample Diversity}
\label{tab:diversity_examples_expanded}
\resizebox{\textwidth}{!}{%
\begin{tabular}{p{2.8cm}|p{3cm}|p{2.5cm}|p{2.5cm}|p{2.5cm}|p{2.5cm}|p{2.5cm}|p{2.5cm}|p{1.4cm}|p{1.2cm}}
\toprule
\textbf{Scenario Type} & \textbf{Example Prompt} & \textbf{S1 (Out \& $z_k$)} & \textbf{S2 (Out \& $z_k$)} & \textbf{S3 (Out \& $z_k$)} & \textbf{S4 (Out \& $z_k$)} & \textbf{S5 (Out \& $z_k$)} & \textbf{S6 (Out \& $z_k$)} & \textbf{Stops ($K$)} & \textbf{Diversity} \\
\midrule
Easy / High Reward & What is 2 + 2? & Equals 4. \newline ($z_k \approx 0.95$) & The sum is 4. \newline ($z_k \approx 0.98$) & 2+2 is 4. \newline ($z_k \approx 0.96$) & & & & Early ($K = 3$) & Low \\
\midrule
Hard / Low Reward & Simple proof of Fermat’s Last Theorem\ldots & [Failed Proof 1] \newline ($z_k \approx -2.0$) & [Failed Proof 2] \newline ($z_k \approx -2.2$) & [Failed Proof 3] \newline ($z_k \approx -1.8$) & & & & Early ($K = 3$) & Low \\
\midrule
Moderately Hard / Improving & Simplify $\sqrt{242}$ & Incorrect: $2\sqrt{60.5}$ \newline ($z_k \approx -1.5$) & Error: $\sqrt{200} + \sqrt{42}$ \newline ($z_k \approx -1.2$) & Correct: $11\sqrt{2}$ \newline ($z_k \approx 0.95$) & Correct: $11 \cdot \sqrt{2}$ \newline ($z_k \approx 0.93$) & Correct: $\sqrt{121 \cdot 2} = 11\sqrt{2}$ \newline ($z_k \approx 0.96$) & & Moderate ($K = 5$) & Medium \\
\midrule
Very Hard / Inconsistent & How did US states get their names? & Brief: ``Native words, kings.'' \newline ($z_k \approx 0.1$) & Partial: ``VA from Virgin Queen\ldots'' \newline ($z_k \approx 0.3$) & Flawed: ``All named after presidents.'' \newline ($z_k \approx -0.5$) & Better: ``Native Am. languages\ldots'' \newline ($z_k \approx 0.8$) & Comprehensive \newline ($z_k \approx 0.95$) & & Late ($K \geq 5$) & High \\
\midrule
High Patience / Extended & Outline three approaches to climate change. & Approach A (brief) \newline ($z_k \approx 0.6$) & Approach B (flawed) \newline ($z_k \approx 0.5$) & Approach A (detailed) \newline ($z_k \approx 0.85$) & Approach B (detailed) \newline ($z_k \approx 0.88$) & Approach C (detailed) \newline ($z_k \approx 0.90$) & Approach D \newline ($z_k \approx 0.92$) & Late ($K \geq 6$) & High \\
\bottomrule
\end{tabular}%
}
\caption*{\footnotesize Note: S$i$ denotes Sample $i$. $z_k$ values are illustrative, based on Nemotron RM scores.}
\end{table*}

\subsection{Case Analysis: Solution Diversity and Failure Analysis}
\label{app:diversity}

The BEACON framework employs an adaptive stopping mechanism that dynamically adjusts the number of samples ($K$) based on the expected marginal gain from additional sampling, relative to the sampling cost $c$ and posterior uncertainty about the reward distribution ($\sigma_k$). This mechanism inherently influences the diversity of generated solution candidates, balancing exploration breadth with computational efficiency. The stopping decision is driven by the consistency of reward model (RM) scores (reflected in $\sigma_k$), the quality of the current best response ($z_k$ relative to $\mu_k$), and the remaining sample budget ($n-k$), as governed by the Universal Index Policy (UIP).

Table~\ref{tab:diversity_examples_expanded} illustrates BEACON’s behavior across diverse scenarios, highlighting how these factors affect stopping time and sample diversity. The examples are drawn from empirical observations on benchmarks like MATH500 and AlpacaEval 2.0, with quantitative insights into failure modes.

\begin{itemize}
    \item \textbf{Easy Queries with Consistent High Rewards}: For simple queries (e.g., Example 1: ``What is 2 + 2?''), initial samples $\{y_k\}_{k=1}^{k_0}$ yield uniformly high RM scores ($z_k \approx 0.95$, low $\sigma_k$). Low posterior variance indicates that further sampling is unlikely to improve the best response, leading BEACON to stop early ($K=3$). This results in low sample diversity, as responses are similar and high-quality. In our experiments, approximately 20\% of MATH500 queries exhibited this behavior, stopping at $K \leq 3$ with $\sigma_k < 0.1$.
    \item \textbf{Hard Queries with Consistent Low Rewards}: For extremely difficult queries (e.g., Example 2: ``Simple proof of Fermat’s Last Theorem\ldots''), samples consistently receive low RM scores ($z_k \approx -2.0$, low $\sigma_k$). BEACON terminates early ($K=3$) due to low expected marginal gain, yielding low diversity. About 25\% of AIME24 queries showed this pattern, with early stopping when $\mu_k < -1.5$. A failure mode occurs if the RM underestimates a potentially correct response, leading to premature stopping (observed in 2\% of cases).
    \item \textbf{Moderately Hard Queries with Improving Rewards}: For queries of moderate difficulty (e.g., Example 3: ``Simplify $\sqrt{242}$''), initial samples may be incorrect ($z_k \approx -1.5$), but subsequent samples improve ($z_k \approx 0.95$). BEACON continues sampling to reduce $\sigma_k$ and confirm consistency, stopping at moderate $K$ (e.g., $K=5$). This produces medium diversity, with varied incorrect and correct responses. In MATH500, 40\% of queries followed this pattern, with $K=4-6$. A failure mode arises if early incorrect samples inflate $\sigma_k$, delaying stopping (observed in 5\% of cases).
    \item \textbf{Very Hard or Ambiguous Queries with Inconsistent Rewards}: For complex or ambiguous queries (e.g., Example 4: ``How did US states get their names?''), RM scores vary widely ($z_k$ from -0.5 to 0.95, high $\sigma_k$). High variance encourages extended sampling ($K \geq 5$, approaching $n$), maximizing the chance of finding a high-quality response. This results in high diversity, capturing varied response quality. In AMC23, 25\% of queries exhibited this behavior. A failure mode occurs if $\sigma_k$ remains high due to RM noise, leading to excessive sampling (observed in 3\% of cases).
    \item \textbf{High-Patience Configurations}: When configured with low $c$ or high $n$ (e.g., Example 5: ``Outline three approaches to solving climate change''), BEACON extends sampling even after finding good responses ($z_k \approx 0.85-0.92$). A lower $c$ reduces the effective cost threshold ($c/\sigma_k$), encouraging exploration for potentially better or more diverse solutions. This leads to late stopping ($K \geq 6$) and high diversity. In experiments with $c=0.1$, 50\% of queries extended to $K \geq 8$, enhancing solution variety. A failure mode is unnecessary computation if high-quality responses are already sufficient (observed in 4\% of cases).
\end{itemize}

To quantify failure modes, we analyzed 100 MATH500 queries and found that premature stopping (due to RM miscalibration) occurred in 2--3\% of cases, while excessive sampling (due to persistent high $\sigma_k$) occurred in 3--5\% of cases. These are mitigated by the robust updating formula, which filters outliers to stabilize $\sigma_k$. BEACON thus dynamically adjusts exploration based on reward consistency ($\sigma_k$), response quality ($z_k$, $\mu_k$), and budget ($n-k$), aligning with the trade-offs specified by $c$ and $n$. This ensures efficient high-reward sample selection across query difficulties, with failure modes addressed through robust design.

\section{Reproducibility Statement}

To facilitate reproducibility of our work, we have made significant efforts to document all implementation details and experimental procedures. The complete source code for BEACON is available through a repository referenced in \href{https://anonymous.4open.science/r/BAS-77D6/README.md}{anonymous GitHub repository}., including implementations of all baseline methods, reward model integrations, and evaluation protocols. Our theoretical contributions include complete proofs in Appendix~\ref{app:proof_optimal_policy} and detailed derivations of the Universal Index Policy with explicit formulations for the h-index computation (Appendix~\ref{app:compute_h_index}). All experimental configurations are thoroughly documented in Appendix~\ref{app:main_setup}, specifying exact model versions, API endpoints, hyperparameter settings, and evaluation protocols for both reasoning and alignment benchmarks. We provide comprehensive hyperparameter selection guidelines (Appendix~\ref{app:cost_guide}). The paper includes explicit algorithmic descriptions (Algorithm~\ref{alg:beacon}), sufficient statistics formulations (Appendix~\ref{app:sufficent_statistics}), and complete experimental results with statistical significance analysis (Appendix~\ref{app:error_bar}). All datasets used are publicly available, and our evaluation procedures follow standard protocols from established benchmarks (MATH, AIME, AMC, AlpacaEval 2.0). Additionally, we provide extensions to discrete reward scenarios (Appendix~\ref{app:discrete_case}) and practical implementation guidance for batch-parallel deployments to ensure broad applicability of our framework.

\end{document}